\documentclass[10pt,twocolumn,letterpaper]{article}

\usepackage[pagenumbers]{cvpr} 

\usepackage{graphicx}
\usepackage{amsmath}
\usepackage{amssymb}
\usepackage{colortbl}
\usepackage{booktabs}
\usepackage{xcolor}         
\usepackage{color}
\usepackage{graphicx}
\usepackage{algorithm}
\usepackage{algorithmicx}
\usepackage{multirow}
\usepackage{multicol}
\usepackage{algpseudocode}
\usepackage{subcaption}
\usepackage{pifont}
\newcommand{\cmark}{\ding{51}}%
\newcommand{\xmark}{\ding{55}}%
\usepackage[pagebackref,breaklinks,colorlinks]{hyperref}
\definecolor{newblue}{HTML}{1770b3}
\hypersetup{colorlinks=true,citecolor={newblue},urlcolor={red}}
\definecolor{blue_new}{rgb}{0.06, 0.75, 0.99}
\definecolor{light_pink}{rgb}{1.0, 0.71. 0.76}
\newcommand{\app}{\raise.17ex\hbox{$\scriptstyle\sim$}}

\usepackage[capitalize]{cleveref}
\crefname{section}{Sec.}{Secs.}
\Crefname{section}{Section}{Sections}
\Crefname{table}{Table}{Tables}
\crefname{table}{Tab.}{Tabs.}

\makeatletter\renewcommand\paragraph{\@startsection{paragraph}{4}{\z@}
	{.5em \@plus1ex \@minus.2ex}{-.5em}{\normalfont\normalsize\bfseries}}\makeatother

\def\cvprPaperID{5386} 
\def\confName{CVPR}
\def\confYear{2023}

\begin{document}

\title{SMAUG: Sparse Masked Autoencoder for Efficient \\ Video-Language Pre-training}

\author{
  Yuanze Lin$^1$ ~~  Chen Wei$^2$ ~~ Huiyu Wang$^2$ ~~ Alan Yuille$^2$ ~~ Cihang Xie$^3$ \vspace{.3em}\\ 
  $^1$ University of Washington ~~ $^2$ Johns Hopkins University ~~
    $^3$ UC Santa Cruz \\
}

\maketitle

\begin{abstract}
Video-language pre-training is crucial for learning powerful multi-modal representation. However, it 
typically requires a massive amount of computation. 
In this paper, we develop SMAUG, an efficient pre-training framework for 
video-language models. 
The foundation component in SMAUG is masked autoencoders.
Different from prior works which only mask textual inputs, our masking strategy considers both visual and textual modalities, providing a better cross-modal alignment and saving more 
pre-training costs. On top of that, we introduce a space-time token sparsification module, which leverages context information to further select only ``important'' spatial regions 
and temporal frames 
for pre-training. 
Coupling all these designs allows our method to enjoy both competitive performances on text-to-video retrieval and video question answering tasks, and much less pre-training costs by \textbf{1.9$\times$} or more. For example, our SMAUG  
only needs \textbf{\app50} NVIDIA A6000 GPU hours for pre-training to attain competitive performances on these two video-language tasks across six popular benchmarks.

\end{abstract}

\section{Introduction}
Recently, video-language pre-training~\cite{lei2021less, li2020hero, bain2021frozen, zhang2021vinvl, fu2021violet, luo2021clip4clip, wang2022all} stands as the 
common practice to learn cross-modal representations on large-scale video-text datasets~\cite{sharma2018conceptual, changpinyo2021conceptual, bain2021frozen}. Such pre-trained models show strong transfer performances on 
a range of vision and language tasks, including visual question answering~\cite{xu2016msr, lin2022revive}, text-to-video retrieval~\cite{xu2016msr, anne2017localizing}, visual reasoning~\cite{suhr2018corpus} and video understanding~\cite{lin2021self, buch2022revisiting, wang2022adafocus}. 
Nonetheless, the corresponding training cost of these advanced video-language models 
is enormous. For example, the training of CLIP4Clip~\cite{luo2021clip4clip} needs \app2 weeks with 8 GPUs, which therefore largely limits their explorations in a wider aspect.
This invites us to ponder 
a thought-provoking but rarely explored question in this paper: \textit{How can we still pre-train powerful video-language models while significantly reducing their pre-training cost?}

\begin{figure}[t]
\centering
\includegraphics[width=1\linewidth]{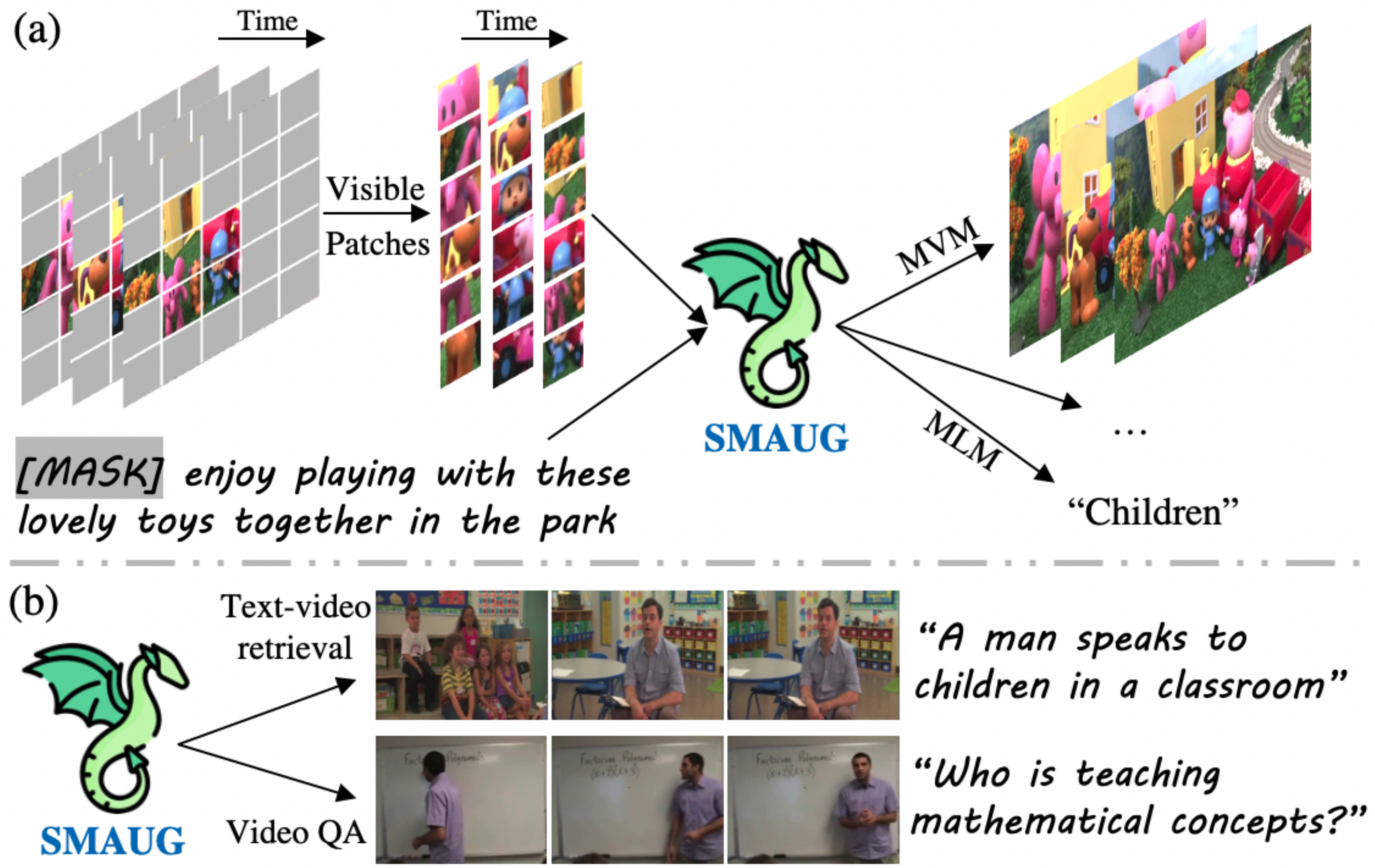}
\vspace{-1.8em}
    \caption{An overview of SMAUG (\textbf{S}parse \textbf{M}asked \textbf{A}utoencoder for video-lang\textbf{U}a\textbf{G}e pre-training). (a) During pre-training, we randomly mask out a large subset of individual frames' patches, and then utilize the \textit{visible patches} and language sentences for video-text pre-training, which includes masked visual/language modeling (MVM/MLM) and \textit{etc}. (b) The pre-trained models can then be fine-tuned on several down-stream video-language tasks, \textit{e.g.}, text-to-video retrieval and video question answering (video QA).}
\label{fig:sample}
\vspace{-1em}
\end{figure}

Interestingly, we note that the recent work Masked Autoencoders (MAE)~\cite{he2022masked}, which establishes an efficient self-supervised paradigm for training models at scale,
potentially offering
a solution to the aforementioned question.
In MAE, a large amount of image patches (\eg, 75\%) are masked. The heavy encoder only executes on a small portion of the visible patches, and the lightweight decoder reconstructs the other large portion of masked patches. This mask reconstruction process is a computationally efficient instantiation of masked visual modeling (\ie, MVM), which has already been shown effective for helping video-language pre-training~\cite{fu2022empirical, kwon2022masked}. 
Therefore we conjecture that resorting to such MAE fashion can substantially mitigate the computational burden and still achieve satisfactory performances for video-language pre-training models.

Another interesting observation is that, even by masking out a significant portion of image patches as in MAE, the information could still be redundant~\cite{wang2022adafocus, feichtenhofer2022masked}. As argued in~\cite{rao2021dynamicvit, liang2022not}, \textit{not all patches are equally important}: an image could contain a significant amount of less informative visual patches (\eg, background patches), which scarcely or even negatively contribute to vision-language representation learning. What is worse, this issue could be severer in the \textit{video}-language setting, as additionally, \textit{not all frames are equally important}~\cite{lei2022revealing, lei2021less}. For example, a video clip may contain a non-negligible portion of frames with just trivial noises (\eg, camera shake). Further eliminating these redundancies is expected to provide an extra speedup to video-language pre-training.

Based on the observations above, we present SMAUG, an efficient pre-training framework for
video-language models. Our work is built upon MAE. We mask out a significant amount of space-time patches and let the autoencoder learn to reconstruct them. 
Next,  we introduce a space-time token sparsification module 
to remove spatial and temporal redundancies: 1) we leverage the attention weights in visual encoder to predict attentive or inattentive tokens to reduce spatial patches among individual frames. Attentive tokens are remained while inattentive tokens are fused; 2) we propose a learnable Transformer-based network to pick up important video frames among the given video clip. 

We evaluate SMAUG on two video-language tasks, including text-to-video retrieval and video question answering across six datasets. For text-to-video retrieval, the experiments are performed on MSRVTT~\cite{xu2016msr}, DiDeMo~\cite{anne2017localizing} and ActivityNet Captions~\cite{krishna2017dense}. For video question answering, MSRVTT-MC~\cite{yu2018joint}, MSRVTT-QA~\cite{xu2017video} and ActivityNet-QA~\cite{yu2019activitynet} are used. SMAUG can achieve state-of-the-art or comparable performances over all six datasets. Meanwhile, the proposed method can achieve \app\textbf{1.9$\times$} video-language pre-training speedup. For example, SMAUG can finish video-language pre-training only with \textbf{\app50} NVIDIA A6000 GPU hours.

\section{Related Work}
\paragraph{Video-and-language pre-training.} The standard pipeline of video-language pre-training (\ie, first pre-train and then fine-tune)~\cite{wang2022all, zhu2020actbert, fu2021violet, luo2021clip4clip, luo2020univl, lei2022revealing} aims at learning a generalizable multi-modal feature representation for a range of downstream tasks, such as text-to-video retrieval~\cite{xu2016msr, anne2017localizing, krishna2017dense, dzabraev2021mdmmt}, video question answering~\cite{yu2018joint, xu2017video, yu2019activitynet, le2020hierarchical}, video captioning~\cite{wang2019vatex, xu2016msr, zhou2018towards, wang2018reconstruction}, \etc. 

UniVL~\cite{luo2020univl} proposes a unified video-language pre-training model for multi-modal generation and understanding. Clip4clip~\cite{luo2021clip4clip} transfers the image-text pre-trained model (\ie, CLIP~\cite{radford2021learning}) for video-retrieval task in an end-to-end manner. Singularity~\cite{lei2022revealing} reveals that the video-language models pre-trained with a single video frame can still attain significant performances for video-and-language downstream tasks. Our work is directly motivated by Singularity~\cite{lei2022revealing}, 
which also pre-trains models by leveraging single-frame and multiple-frame setups.

\paragraph{Masked visual modeling.} The main goal of masked visual modeling (MVM) is to acquire effective visual representations by the first masking and then reconstructing process. The pioneering work, denoising autoencoders (DAE)~\cite{vincent2008extracting, vincent2010stacked}, learns representations by reconstructing the corrupted signals. 
Recently, iGPT~\cite{chen2020generative} regards the pixels as tokens and predicts unknown pixels. MAE~\cite{he2022masked} masks out a subset of patches and learns to predict their original pixels. Other considerations of prediction targets include features~\cite{wei2022masked} and discrete visual tokens~\cite{bao2021beit}.

In addition to image recognition, a set of works~\cite{feichtenhofer2022masked, tong2022videomae, wang2022bevt} further extend MAE into video recognition. In this paper, we focus on exploring the potential of MAE in enabling efficient vision-language pre-training.

\begin{figure*}[t]
\centering
\includegraphics[width=.99\linewidth]{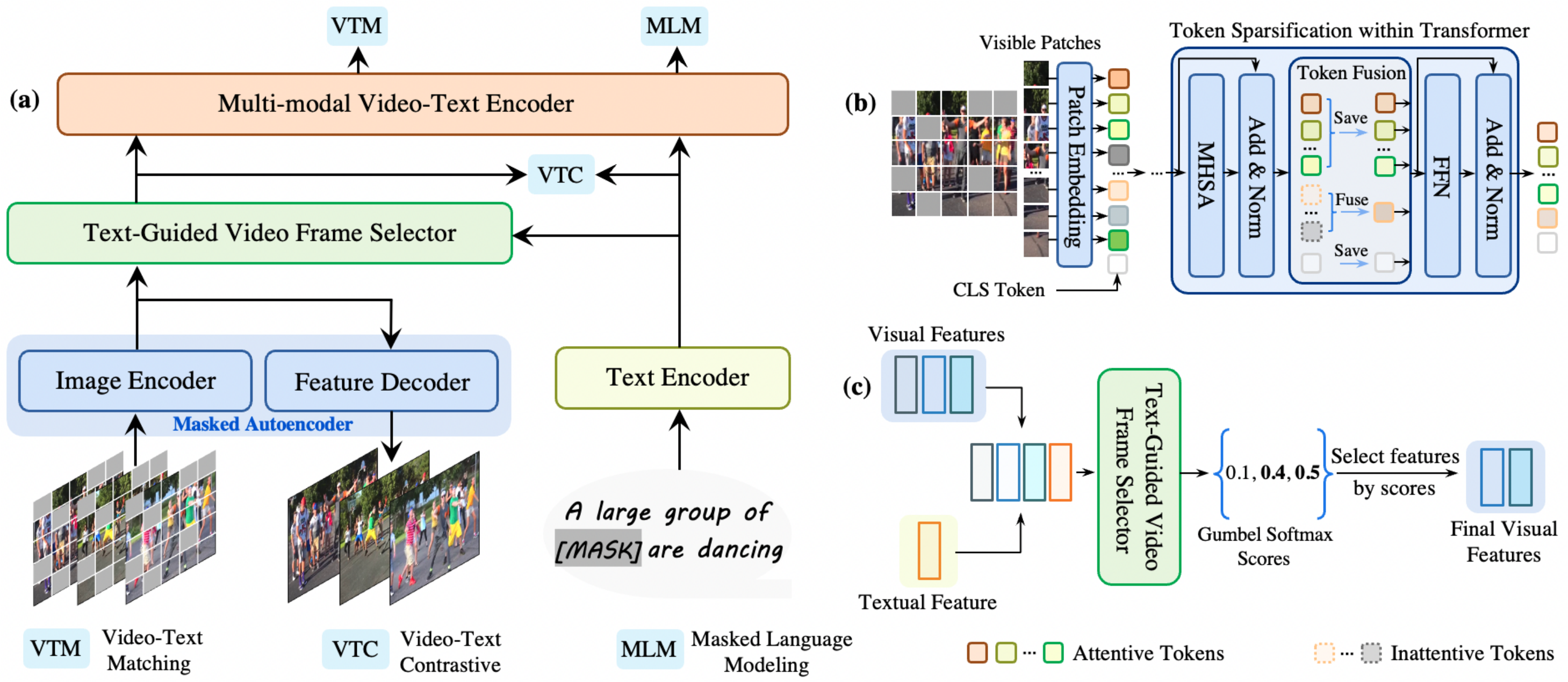}
\vspace{-0.5mm}
    \caption{\textbf{An illustration of SMAUG architecture.} We use three video frames as an example. (a) The pre-training framework of our proposed SMAUG, we adopt MAE to extract visual features from \textit{visible patches} and reconstruct the pixel values of masked patches. Note that we perform visual token sparsification (\textit{i.e.}, (b)) in the image encoder, and the reconstruction loss for masked patches, \textit{i.e.}, masked visual modeling (MVM), is performed in masked autoencoders. (b) The visual token sparsification module reduces spatial redundancies for \textit{visible patches} of individual video frames, we only utilize single frame input for an explanation. (c) Text-guided video frame selector network takes visual and textual features as inputs and outputs the selected frames by the scores. It can further perform sparsification for video frames along the temporal dimension. The example of selecting two frames among the given video is used for explanation.}
\vspace{-1em}
\label{fig:architecture}
\end{figure*}

\paragraph{Efficient vision transformer.} Recent works have started investigating eliminating redundant computations in 
 (cumbersome) Vision Transformer (ViT). DynamicViT~\cite{rao2021dynamicvit} designs a lightweight prediction module first to estimate the importance scores of tokens, and then utilize the scores to prune redundant tokens hierarchically. EVIT~\cite{liang2022not} proposes to reorganize tokens by the attention from the class token, \ie, they preserve informative tokens while fusing uninformative ones. Motivated by EVIT~\cite{liang2022not}, we also leverage attention from the class token to reduce spatial redundancies.

\paragraph{Temporal redundancy.} Giving not all video frames equally contribute to video recognition, many works have been proposed to reduce the temporal redundancies among videos~\cite{wang2022adafocus, gao2020listen, ghodrati2021frameexit, kim2021efficient, korbar2019scsampler, meng2020ar, sun2021dynamic}. AR-Net~\cite{meng2020ar} selects the optimal frame resolution conditioned on inputs for efficient video recognition by a decision policy. FrameExit~\cite{ghodrati2021frameexit} can automatically choose the number of frames to process, according to the complexities of the given videos. This paper proposes a Transformer-based network conditioned on the given video and language features to select informative video frames for reducing temporal redundancies.  

\section{Method}
We first introduce the pre-training model architecture in Section \ref{architecture}. Second, we elaborate on the modules to reduce spatial and temporal redundancies in Section \ref{redundancy}. Third, we describe the pre-training objectives in Section \ref{objectives}. Fourth, we list the pre-training datasets in Section \ref{data} and implementation details in Section \ref{implementation}.

\paragraph{Overview.} Given a video clip $V=[v_{1}, v_{2}, ..., v_{N}]$ with $N$ video frames and a text sentence $L$, the video-language pre-training model needs to extract the visual and textual embeddings separately and feed the embeddings into the multimodal encoder to learn cross-modal representations. A decoder can optionally process the learned cross-modal representations to generate final outputs. 

\subsection{Model Architecture}
\label{architecture}
As illustrated in Figure \ref{fig:architecture}, the proposed SMAUG consists of two major components: \textit{a}) a video-language pre-training model and \textit{b}) space-time redundancy sparsification modules. The video-language pre-training model mainly contains \textit{a}) a masked autoencoder, \textit{b}) a text encoder, and \textit{c}) a multi-modal video-text encoder for cross-modal representation fusion. The space-time redundancy sparsification modules aim to further reduce spatial and temporal redundancies among the visible patches.

\subsubsection{Masked Autoencoder}
We introduce MAE~\cite{he2022masked} into video-language pre-training. MAE contains an encoder $\mathcal{Q}_{v}$ and a decoder $\mathcal{D}_{v}$. We explain the involved components as follows:

\paragraph{Frame sampling.} Following the setup of Singularity~\cite{lei2022revealing}, we adopt both single-frame and multi-frame options for pre-training. Specifically, we randomly sample one video frame or multiple video frames from the given video clip $V$ to pre-train the video-language models.

\paragraph{Patch embedding.} Following ViT~\cite{dosovitskiy2020image}, we first patchify the sampled frames into non-overlapping patches~\cite{tong2022videomae, feichtenhofer2022masked}, which are then flattened and projected by a linear layer. The position embeddings~\cite{vaswani2017attention} are added into the projected patches. In particular, the encoder (\ie, ViT) of MAE~\cite{he2022masked} operates independently on each frame.

\paragraph{Masking.} We randomly mask out a subset of embedded patches for each frame and use the remaining \textit{visible} ones. Note that we perform tube masking~\cite{wei2022masked} for multi-frame input, \ie, the spatial locations of the masked patches are the same for all sampled frames. Although videos contain abundant space-time redundancies, masking out too many patches could lead to inaccurate alignment between visual and textual representations. In our experiments, we at most set the mask ratio to be 65\%.

\paragraph{Feature encoding.} The encoder $Q_{v}$ operates only on the visible embedded patches and omits the masked ones following MAE~\cite{he2022masked}. The output of the encoder $\mathcal{F}_{v}$ is a sequence of visual embeddings. When only single-frame input is sampled for pre-training, the output $\mathcal{F}_{v}$ can be represented as:
\begin{equation}
    \label{visual}
    \mathcal{F}_{v} = \{f_{\text{cls}}, f_{\text{1}}, ..., f_{L}\}, 
\end{equation}
in which $f_{i} \in \mathbb{R}^{d}$, $L$ is the sequence length of visible patches, and $d$ denotes the feature dimension. $f_{\text{cls}}$ represents the visual \texttt{[CLS]} token. When sampling multiple frames for pre-training, we additionally adopt a temporal encoder $\mathcal{Q}_{t}$ to perform self-attention on frame-wise features across the temporal dimension. The output $\mathcal{F}_{v}$ now is:
\begin{equation}
    \label{visual2}
    \mathcal{F}^{k}_{v} = \{f^{k}_{\text{cls}}, f^{k}_{\text{1}}, ..., f^{k}_{L}\}, \quad k=1, ..., K, 
\end{equation}
\begin{equation}
    \label{visual3}
    \mathcal{F}_{v} = \mathcal{Q}_{t}(\text{Concat}(\mathcal{F}^{\text{1}}_{v}, ...,\mathcal{F}^{K}_{v})),
\end{equation}
where $K$ is the number of sampled frames.
$\mathcal{F}_{v}$ are adopted for cross-modal alignment.

\paragraph{Feature decoding.} After that, we feed the embeddings of \textit{visible patches} and the masked patches together into the decoder $D_{v}$ for predicting the pixels of these masked ones. This process can be regarded as masked visual modeling (MVM), which has been adopted as one of the pre-training objectives. There exists other reconstruction targets \cite{fu2022empirical} except for pixels, \eg, features, depth maps, \textit{etc}. We leave them in the future.

\subsubsection{Text Encoder and Multi-modal Fusion}
We then describe the details of the other encoders for textual embedding extraction and cross-modal fusion.

\paragraph{Text encoder.} Given the text sentence $L$, the text encoder $\mathcal{Q}_{l}$ firstly segments it into a sequence of subwords~\cite{sennrich2015neural}, and inserts the special token (\eg, \texttt{[CLS]} token) at the beginning of the subword sequence. This token sequence is then mapped into the text embeddings $\mathcal{F}_{l}$, which can be represented as:
\begin{equation}
    \label{text}
    \mathcal{F}_{l} = \{h_{\text{cls}}, h_{\text{1}}, ..., h_{P}\}, 
\end{equation}
in which $h_{i} \in \mathbb{R}^{d}$, $h_{\text{cls}}$ is the text embedding of text \texttt{[CLS]} token and $P$ means the number of text tokens.

\paragraph{Multi-modal video-text encoder.} After obtaining the visual embedding $\mathcal{F}_{v}$ and text embeddings $\mathcal{F}_{l}$, we feed them into the multi-modal encoder $\mathcal{Q}_{m}$ to perform multi-modal fusion by cross-attention layers \cite{lei2021less,lei2022revealing,jiang2022pseudo}. The learned cross-modal representations are used for video-text matching and masked language modeling objectives (\textit{i.e.}, VTM and MLM) to pre-train the models.

\subsection{Space-Time Token Sparsification}
\label{redundancy}
Although MAE has already randomly removed a large subset of frame patches, there still exist spatial and temporal redundancies among the remaining \textit{visible patches} \cite{rao2021dynamicvit,liang2022not,wang2022adafocus,ghodrati2021frameexit}, because MAE masks patches randomly without considering how informative they are. In order to make an informative decision, we introduce two context-dependent selection modules, for keeping informative tokens and removing uninformative ones. These modules can further save a large amount of pre-training costs (Table \ref{table_a4}).

\paragraph{Visual token sparsification.} A considerable amount of uninformative visual tokens (\eg, background patches) will have little impact on the final performance when they are removed~\cite{liang2022not}. Since the MAE encoder $\mathcal{Q}_{v}$ inherits the structure of ViT, we can insert a token sparsification module, inspired by EVIT~\cite{liang2022not}, in the 4$^{th}$, 7$^{th}$ and 10$^{th}$ layers of the ViT-B encoder $\mathcal{Q}_{v}$. The mechanism of token sparsification within a transformer layer is shown in Figure~\ref{fig:architecture}(b).

Specifically, the token sparsification module computes the average attention values among all heads from \texttt{[CLS]} with respect to the other tokens. The tokens whose corresponding attention values are top-$k$ largest are regarded as attentive tokens, otherwise inattentive tokens. We retain the attentive tokens while fusing the inattentive tokens into a new token by taking the attention values as coefficients. The keeping rate of the tokens is defined as $\gamma=k/p$, where $k$ and $p$ mean the number of attentive tokens and total input tokens, and the attentive tokens have top $\gamma*100\%$ average attention values. We explain the details in the Appendix.

\paragraph{Text-guided video frame selection.} Similar to the image-level redundancy, videos, as collections of frames, are also redundant on the temporal axis. A fraction of frames could contain more informative contexts than others especially when corresponding text sentences are given. We hereby aim to select the most context-relevant frames from a given clip to perform video-language tasks.

To this end, we propose a Transformer-based frame selector $\mathcal{S}$, which is illustrated in Figure~\ref{fig:architecture}(c). Specifically, it's exemplified by the case of selecting the features of top-2 essential frames. It first takes the visual and textual embeddings from Equation (\ref{visual3}) and (\ref{text}) respectively, and then outputs the features of essential frames. The frame selection process is denoted as:
\begin{equation}
\begin{split}
    \label{select}
    \mathcal{I} &= S(\mathcal{F}_{v}; \mathcal{F}_{l}) \\
    &= \{\mathcal{F}^{\mu_{\text{1}}}_{v}, ..., \mathcal{F}^{\mu_{\kappa}}_{v}\}, \quad i = 1, ..., \kappa.
\end{split}
\end{equation}

In Equation (\ref{select}), $\mathcal{I} \in \mathbb{R}^{\kappa \times L^{\prime} \times d}$ ($1<=\kappa<K$) denotes the sparse collection of video frame features selected from the original visual embeddings $\mathcal{F}_{v}$, and $L^{\prime}$ is the sequence length of embedded patch embeddings. Since the frame selection process is discrete, we adopt the Gumbel-Softmax trick~\cite{jang2016categorical} to optimize the parameters of selector $\mathcal{S}$ in both the full pre-training stage and the task-specific fine-tuning stage. At inference time, the operation is discrete by selecting the top-k ($k=\kappa$) features as the final output. In this way, we obtain a sparse and efficient model.

\subsection{Pre-training Objectives}
\label{objectives}
Our models are pre-trained by using four objectives: (1) Video-Text Matching ($\mathcal{L}_{\text{vtm}}$): it predicts the matching scores of the given video-text pairs through the multi-modal video-text encoder's outputs. (2) Masked Language Modeling ($\mathcal{L}_{\text{mlm}}$): it fuses both visual and textual features by the multi-modal video-text encoder to predict the masked textual tokens. (3) Video-Text Contrastive ($\mathcal{L}_{\text{vtc}}$): it uses the pooled visual and textual representations to align parallel video-text pairs, so that they can have higher similarity scores. (4) Masked Visual Modeling ($\mathcal{L}_{\text{mvm}}$): it learns to predict the original pixels of masked visual patches.  

These objectives have been widely used for pre-training video-language models~\cite{lei2022revealing, fu2021violet, wang2022all, fu2022empirical, li2022align, dou2022empirical}, we describe the details of them in the Appendix. The full pre-training objective $\mathcal{L}$ is the simple addition of these four objectives:
\begin{equation}
    \label{loss}
    \mathcal{L} = \mathcal{L}_{\text{vtm}} + \mathcal{L}_{\text{mlm}} + \mathcal{L}_{\text{vtc}} + \mathcal{L}_{\text{mvm}}.
\end{equation}

\subsection{Pre-training Datasets}
\label{data}
We adopt video-text and image-text data for pre-training. WebVid~\cite{bain2021frozen} is used as video-text data, which scrapes 2.5M video-text pairs from the web. The image-text data includes the combination of CC3M~\cite{sharma2018conceptual}, CC12M~\cite{changpinyo2021conceptual}, SBU Captions~\cite{ordonez2011im2text}, Visual Genome (VG)~\cite{krishna2017visual} and COCO~\cite{lin2014microsoft} datasets. Two subsets are employed for pre-training: 1) 5M corpus consisting of CC3M and WebVid. 2) 17M corpus that contains all the mentioned image-text and video-text datasets. Note that when using the single frame setting, we also sample only one video frame from the video-text dataset, \ie, WebVid.

\subsection{Implementation Details}
\label{implementation}

\paragraph{Network structures.} For the masked autoencoder, we adopt ViT-B~\cite{dosovitskiy2020image} as the encoder $\mathcal{Q}_{v}$, whose weights are initialized from CLIP's visual encoder (\ie, CLIP-B/16)~\cite{radford2021learning}. The decoder $\mathcal{D}_{v}$ is the stack of one linear layer, three Transformer blocks, and one linear layer. We adopt BERT$_{BASE}$ model as our text encoder $\mathcal{Q}_{l}$ and multi-modal video-text encoder $\mathcal{Q}_{m}$. In particular, the first 9 layers and the last three layers of BERT$_{BASE}$ are initialized as $\mathcal{Q}_{l}$  and $\mathcal{Q}_{m}$, respectively. The cross-attention layers \cite{lei2021less,lei2022revealing} of the multimodal encoder are learned from scratch. The frame selector $\mathcal{S}$ consists of two linear layers and two Transformer blocks~\cite{vaswani2017attention}. The Transformer blocks have two heads and a hidden size of 2048.

\paragraph{Model pre-training.} We use 4$\times$NVIDIA A6000 GPUs to pre-train our models with AdamW optimizer~\cite{loshchilov2017decoupled} and an initial learning rate of $\rm1e^{-4}$. We randomly sample one frame for the single-frame setting and four frames for the multiple-frame setting. The total pre-training epochs are 10, and the learning rate is warmed up in the first epoch, followed by cosine decay~\cite{loshchilov2016sgdr} to $\rm1e^{-6}$ finally. The input size of the video frames is 224$\times$224. The data augmentation includes random resized crop and flip operations. When using the single frame for pre-training, our model only takes about 13 hours to pre-train on the 5M corpus and two days on the 17M corpus, attaining better performance than ALPRO~\cite{li2022align}, which takes three days to pre-train the same length schedule epochs (\ie, 10 epochs) on the 5M corpus with 16$\times$A100 GPUs. It is worth mentioning that when using multiple video frames, \textit{i.e.}, four frames, for pre-training, the model weights are initialized from single-frame pre-trained models, and the total training epoch for such setup is 5.

\section{Experiment}

\begin{table*}[htp]
{\centering}
\resizebox{1\textwidth}{!}{%
\begin{tabular}{cccccccccccc}
\multicolumn{1}{c}{\multirow{2}{*}{Method}} & \multicolumn{1}{c}{\multirow{2}{*}{PT Datasets}} 
& \multicolumn{1}{c}{\multirow{2}{*}{\#Frame}} 
&\multicolumn{3}{c}{MSRVTT} & \multicolumn{3}{c}{DiDeMo} & \multicolumn{3}{c}{ActivityNet Cap} %
\\ 
\multicolumn{1}{c}{} & \multicolumn{1}{c}{} & \multicolumn{1}{c}{} & R@1 & R@5 & R@10 & R@1 & R@5 & R@10 & R@1 & R@5 & R@10 \\ 
\midrule
\textcolor{black}{\textit{Pre-trained with $>$100M video-text pairs}} \\
HT100M~\cite{miech2019howto100m} & HT100M & 16 & 14.9 & 40.2 & 52.8 & - & - & - & - & - & - \\
\rowcolor{light_pink!15} HERO~\cite{li2020hero} & HT100M  & 310 & 20.5 & 47.6 & 60.9 & - & - & - & - & - & -  \\
\rowcolor{light_pink!15} MMT~\cite{gabeur2020multi} & HT100M& 1K/-/3K & 26.6 & 57.1 & 69.6 & - & - & - & 28.7 & 61.4 & 94.5 \\
\rowcolor{light_pink!15} AVLNet~\cite{rouditchenko2020avlnet} & HT100M & - & 27.1 & 55.6 & 66.6 & - & - & - & - & - & - \\
SupportSet~\cite{patrick2020support} & HT100M & - &  \color{gray}{30.1} & \color{gray}{58.5} & \color{gray}{69.3} & - & - & - & - & - & - \\ 
VideoCLIP~\cite{xu2021videoclip} & HT100M & 960 & 30.9 & 55.4 & 66.8 & - & - & - & - & - & - \\
VIOLET~\cite{fu2021violet} &  YT180M+5M & 4 & 34.5 & 63.0 & 73.4 & 32.6 & 62.8 & 74.7 & - & - & - \\
All-in-one~\cite{wang2022all} &  HT100M+WebVid & 9 & 34.4 & 65.4 & 75.8 & 32.7 & 61.4 & 73.5 & 22.4 & 53.7 & 67.7 \\
\midrule
\textit{Pre-trained with $<$100M video-text pairs} \\
ClipBERT~\cite{lei2021less} & COCO + VG  & 16/16/8 & 22.0 & 46.8 & 59.9 & 20.4 & 48.0 & 60.8 & 21.3 & 49.0 & 63.5  \\
Frozen~\cite{bain2021frozen} &  5M & 4 & \color{gray}{31.0} & \color{gray}{59.5} & \color{gray}{70.5} & 31.0 & 59.8 & 72.4 & - & - & - \\
ALPRO~\cite{li2022align} &  5M  & 8 & 33.9 & 60.7 & 73.2 & 35.9 & 67.5 & 78.8 & - & - & - \\
Singularity~\cite{lei2022revealing} & 5M  & 1 & 36.8 & 65.9 & 75.5 & 47.4 & 75.2 & 84.0 & 43.0 & 70.6 & 81.3 \\
Singularity~\cite{lei2022revealing} & 17M  & 1 & 41.5 & 68.7 & 77.0 & 53.9 & 79.4 & 86.9 & 47.1 & 75.5 & 85.5 \\
\midrule
\textbf{Ours} & \textbf{5M}  & \textbf{1} & \textbf{40.6} & \textbf{67.6} & \textbf{77.5} & \textbf{49.2} & \textbf{76.7} & \textbf{85.6} & \textbf{44.8}  & \textbf{72.2}  &  \textbf{82.7} \\
\textbf{Ours} & \textbf{17M}  & \textbf{1} & \textbf{44.0} & \textbf{70.4} & \textbf{78.8} & \textbf{55.6} & \textbf{80.8}  & \textbf{88.4}  &  \textbf{49.2} & \textbf{76.9}  & \textbf{86.8}  \\
\label{table1}
\end{tabular}
}
\vspace{-5mm}
\caption{
Fine-tuning results compared with existing video-language pre-training methods on text-to-retrieval. The pre-training (PT) text-video datasets are HowTo100M (HT100M)~\cite{miech2019howto100m}, YT-Temporal-180M (YT180M)\cite{zellers2021merlot}, MS-COCO (COCO)~\cite{lin2014microsoft}, Visual Genome (VG)~\cite{krishna2017visual}, 5M corpus and 17M corpus. Note that 5M and 17M settings are illustrated in Section \ref{data}. For MSRVTT dataset, results using 9K training videos are \textbf{{\color{gray}{greyed out}}}. The methods using other modalities (\eg, speech and audio) are highlighted by \colorbox{light_pink!30}{pink}. ``\#Frame" denotes the number of final video frames for video-language model pre-training. For those methods which adopt different input frames for different datasets (\ie, MMT~\cite{gabeur2020multi} and ClipBERT~\cite{lei2021less}), we leverage ``/" to separate them.} 
\label{table1}
\vspace{-2mm}
\end{table*}

\subsection{Down-Stream Tasks and Datasets}
\paragraph{Text-to-video retrieval.} We first evaluate our approach on text-to-video retrieval across three video-language datasets.  The recall performance at K (R@K) is reported. 
\begin{itemize}
\item \textbf{MSRVTT}~\cite{xu2016msr} consists of 10K videos from YouTube, and each video has 20 textual captions. We follow the standard setting as~\cite{lei2022revealing,li2022align,miech2019howto100m,yu2018joint}, \ie, we fine-tune the pre-trained models with 7K videos and report the performances on the 1K test split~\cite{yu2018joint}.

\item \textbf{DiDeMo}~\cite{anne2017localizing} includes 10K videos, which are collected from Flickr; there are 41K text descriptions total, and the train/val/test splits are adopted. 

\item  \textbf{ActivityNet Captions}~\cite{krishna2017dense} consists of 20K Youtube videos which are paired with 100K captions. We evaluate the results on \textit{val1} split. 
\end{itemize}

For MSRVTT, we report the results of the standard text-to-video retrieval protocol, as for the other two datasets, we perform paragraph-to-video retrieval evaluation~\cite{lei2022revealing,li2022align}, \ie, we concatenate all the captions in the same video to one single paragraph for retrieval.

\paragraph{Video question answering.} We also select video question answering for evaluation, the accuracies are reported, and three benchmarks are chosen: 
\begin{itemize}
\item \textbf{MSRVTT-MC}~\cite{yu2018joint} is a dataset for multiple-choice question answering, which consists of 3K videos and needs to select the best matching caption choice from 5 candidates for each video. 

\item \textbf{MSRVTT-QA}~\cite{xu2017video} is built on MSRVTT, consisting of 10K videos paired with 244K open-ended questions. 

\item \textbf{ActivityNet-QA}~\cite{yu2019activitynet} collects 5.8K videos from ActivityNet~\cite{caba2015activitynet} with 58K open-ended questions.
\end{itemize}

\paragraph{Fine-tuning setups.} When fine-tuning the models on text-to-video retrieval task, we use the same model architecture except that we remove MLM and MVM objectives. The models are trained with an initial learning rate of $\rm1e^{-5}$, which is decreased to $\rm1e^{-6}$ by cosine decay; the training epochs are 5, 10 and 10 for MSRVTT, DiDeMo, and ActivityNet-Captions datasets. During the inference stage, we sample 12 frames from each video for MSRVTT and DiDeMo datasets and 32 frames per video for the ActivityNet Captions dataset.

To evaluate the performances on open-ended video question answering datasets (\eg, MSRVTT-QA and ActivityNet-QA), an extra decoder is added after the multi-encoder, which generates the answers by taking the outputs from the multi-modal encoder. The models are fine-tuned with 10 epochs and tested with 12 frames. As for MSRVTT-MC, we follow the protocol of~\cite{lei2021less}. Specifically, the models trained on MSRVTT is leveraged to select the best matching choice with the highest retrieval scores as final predictions. Note that for all downstream tasks, we resize the video frames as 224$\times$224, and the data augmentation is the same as the pre-training process.

\subsection{Comparison with State-of-the-arts}

Here, we show the performances of the proposed approach, the default masking ratios of MAE and masked language modeling (MLM) are $50\%$ and 15$\%$, and the keeping rate $\gamma$ is 0.8. The number of input frames keeps unchanged during the pre-training and fine-tuning stages.

\begin{table}[t]
{\centering}
\resizebox{0.48\textwidth}{!}{
\begin{tabular}{p{2.5cm}<{\centering}cccccc}
\multicolumn{1}{c}{\multirow{2}{*}{Method}} &\multicolumn{3}{c}{MSRVTT} & \multicolumn{3}{c}{DiDeMo} \\
\multicolumn{1}{c}{} & R@1 & R@5 & R@10 & R@1 & R@5 & R@10\\  
\midrule
\rowcolor{light_pink!15} MMT~\cite{gabeur2020multi} & - & 6.9 & - & - & - & - \\
\rowcolor{light_pink!15} AVLNet~\cite{rouditchenko2020avlnet}  & 19.6 & 40.8 & 50.7 & - & - & - \\
VideoCLIP~\cite{xu2021videoclip} & 10.4 & 22.2 & 30.0 & 16.6 & 46.9 & -  \\
Frozen~\cite{bain2021frozen} & 18.7 & 39.5 & 51.6 & 21.1 & 46.0 & 56.2  \\
ALPRO~\cite{li2022align} & 24.1 & 44.7 & 55.4 & 23.8 & 47.3 & 57.9  \\
VIOLET~\cite{fu2021violet} & 25.9 & 49.5 & 59.7 & 23.5 & 49.8 & 59.8  \\
Singularity~\cite{lei2022revealing} & 28.4 & 50.2 & 59.5 & 36.9 & 61.1 & 69.3 \\
\midrule
\textbf{Ours} & \textbf{28.9} & \textbf{52.1} & \textbf{62.9} & \textbf{34.3} & \textbf{60.3}  & \textbf{69.4} \\
\end{tabular}
}
\caption{Zero-shot results compared with existing video-language pre-training methods on text-to-retrieval. The pre-training text-video datasets and the number of input video frames are the same as Table \ref{table1}. Note that the methods using other modalities (\eg, speech and audio) are highlighted by \colorbox{light_pink!30}{pink}. Note that for a clear comparison, we display the results of Singurity~\cite{lei2022revealing} and our SMAUG, whose number of the final video frames is 1, and the pre-training corpus is 5M.}
\label{table2}
\vspace{-6mm}
\end{table}

\paragraph{Text-to-video retrieval.} In Table \ref{table1}, we show the compared results on text-to-video retrieval task. When the final video frame number is only 1 and using a smaller scale of pre-trained data (\eg, 5M and 17M), SMAUG can surpass existing methods by a large margin. Especially, our method can achieve significant performance gains even compared with other approaches (\eg, HERO~\cite{li2020hero}, MMT~\cite{gabeur2020multi} and AVLNet~\cite{rouditchenko2020avlnet}) that adopt other modalities, such as audio, speech, \etc.

\begin{table*}[t]
\resizebox{\textwidth}{!}{%
\centering
\begin{tabular}{p{6cm}<{\centering}p{2.5cm}<{\centering}p{2cm}<{\centering}p{2.5cm}<{\centering}p{2.5cm}<{\centering}p{2.5cm}<{\centering}}
Method & PT Datasets & \#Frame & MSRVTT-QA & ActivityNet-QA & MSRVTT-MC \\
\midrule
\textcolor{black}{\textit{Pre-trained with $>$100M video-text pairs}} \\
JustAsk~\cite{yang2021just} & HT69M & 640 & 41.5 & 38.9 & - \\
MERLOT~\cite{zellers2021merlot} & YT180M & 5 & 43.1 & 41.4 & 90.9 \\
VideoCLIP~\cite{xu2021videoclip} & HT100M & 960 & - & - & 92.1 \\
VIOLET~\cite{fu2021violet} & YT180M+5M & 4 & 43.9 & - & 91.9 \\
\midrule
\textcolor{black}{\textit{Pre-trained with $<$100M video-text pairs}} \\
ClipBERT~\cite{lei2021less} & COCO + VG & 16 & 37.4 & - & 88.2 \\
ALPRO~\cite{li2022align} & 5M & 16 & 42.1 & - & - \\
Singularity~\cite{lei2022revealing} & 5M & 1 & 42.7 & 41.8 & 92.0\\
Singularity~\cite{lei2022revealing} & 17M & 1 & 43.5 & 43.1 & 92.1 \\
\midrule
\textbf{Ours} & \textbf{5M} & \textbf{1} & \textbf{43.4} & \textbf{42.7} &  \textbf{92.7} \\
\textbf{Ours} & \textbf{17M} & \textbf{1} & \textbf{44.5} & \textbf{44.2} & \textbf{92.9} \\
\end{tabular}
}
\vspace{-2mm}
\caption{Results compared with existing video-language pre-training methods on video question answering task. The pre-training (PT) text-video datasets are HowTo100M (HT100M)~\cite{miech2019howto100m}, YT-Temporal-180M (YT180M)\cite{zellers2021merlot}, MS-COCO (COCO)~\cite{lin2014microsoft}, Visual Genome (VG)~\cite{krishna2017visual}, HowToVQA69M (HT69M)~\cite{yang2021just}, 5M corpus and 17M corpus. Note that 5M and 17M settings are illustrated in Section \ref{data}. ``\#Frame" denotes the number of final video frames for video-language model pre-training.}
\vspace{-5mm}
\label{table3}
\end{table*}

As for MSRVTT dataset, when the pre-training corpus is 5M and the number of the final input frame is 1, our pre-trained model can achieve \textbf{3.8\%} relative improvement on R@1 compared with Singularity~\cite{lei2022revealing}, when pre-trained on the 17M corpus, our method can also lead to \textbf{2.5\%} performance gain om R@1 compared with Singularity. These results can demonstrate the effectiveness of our approach. Note that our pre-training is also much faster compared with Singularity~\cite{lei2022revealing}. The analysis of pre-training costs is shown in Section \ref{analysis}.

 \begin{table}[t]
\resizebox{0.47\textwidth}{!}{
        \begin{tabular}{p{2.5cm}<{\centering}p{2cm}<{\centering}ccc}
        \multicolumn{1}{c}{\multirow{2}{*}{Masking Ratio}} & \multicolumn{1}{c}{\multirow{2}{*}{PT Time}} &\multicolumn{3}{c}{MSRVTT} \\
        \multicolumn{1}{c}{} & \multicolumn{1}{c}{} & R@1 & R@5 & R@10 \\  
        \midrule
        10\% & 70.8 hours & \textbf{40.5} & \textbf{66.3} & \textbf{76.5} \\
        25\% & 64.6 hours & 40.1 & 65.7 & 76.0 \\
        \rowcolor{blue_new!15} 50\%  & 50.4 hours& 39.3 & 65.4 & 75.6 \\
        65\% & 44.3 hours & 38.1 & 64.2 & 75.0 
        \end{tabular}}
        \caption{Ablation study on using different masking ratios.}
        \vspace{-3mm}
          \label{table_a1}
\end{table}

In Table \ref{table2}, we show the zero-shot results to compare our approach with existing methods. Our SMAUG can achieve comparable performances compared with Singularity~\cite{lei2022revealing}. The results can demonstrate that even with \textbf{50\%} mask ratios for masked autoencoder, the pre-trained model can still learn meaningful cross-modal representations for video-language tasks.

\paragraph{Video question answering.} We display the comparison of the results between ours and existing methods in Table \ref{table3} for video question answering. When using 5M and 17M settings for pre-training, our proposed approach can still perform better than Singularity~\cite{lei2022revealing}.

We can also observe that, when pre-training the models on the 17M corpus, our method can still surpass the performances of VIOLET~\cite{fu2021violet} on MSRVTT-QA and MSRVTT-MC benchmarks, note that we use less video-text data for pre-training (17M \textit{v.s.} YT180M+5M).

\subsection{Ablation Study}
In this section, we ablate out the influences of the proposed components. The pre-training corpus is 5M. We report the fine-tuning results on the text-video-retrieval task, and the MSRVTT dataset is chosen. Except for the ablation study of frame selection (\ie, Table \ref{table_a3}), the number of the input frames is 1 for the experiments. \textit{``PT Time"} denotes the total time for model pre-training, we report the GPU hours of using a single A6000 GPU. The results highlighted in \colorbox{blue_new!20}{blue} are used to specify the default settings.

\begin{table}[t]
\resizebox{0.47\textwidth}{!}{%
\centering
        \begin{tabular}{p{2.5cm}<{\centering}p{2cm}<{\centering}ccc}
        \multicolumn{1}{c}{\multirow{2}{*}{Keeping Rate}} & \multicolumn{1}{c}{\multirow{2}{*}{PT Time}} &\multicolumn{3}{c}{MSRVTT} \\
        \multicolumn{1}{c}{} & \multicolumn{1}{c}{} & R@1 & R@5 & R@10 \\  
            \midrule
            0.6 & 44.8 hours & 37.5  & 64.2 & 74.6 \\
            0.7 & 47.6 hours & 38.2 & 64.7 & 75.3 \\
            \rowcolor{blue_new!15} 0.8 &  50.4 hours & 39.3 & 65.4 & 75.6 \\
            0.9 & 53.8 hours & \textbf{39.8} & \textbf{65.9} & \textbf{76.0}  \\
        \end{tabular}
}
        \caption{Ablation study on using different keeping rates.}
        \vspace{-2mm}
          \label{table_a2}
\end{table}

\begin{table}[t]
\resizebox{0.47\textwidth}{!}{%
\centering
      \centering
        \begin{tabular}{p{3.5cm}<{\centering}cccc}
        \multicolumn{1}{c}{\multirow{2}{*}{Frame Selection}} & \multicolumn{1}{c}{\multirow{2}{*}{PT Time}} &\multicolumn{3}{c}{MSRVTT} \\
        \multicolumn{1}{c}{} & \multicolumn{1}{c}{} & R@1 & R@5 & R@10 \\ 
        \midrule
         \textit{Single-frame selection} \\
            1  & \textbf{50.4 hours} & 39.3 & 65.4 & 75.6 \\
            \rowcolor{blue_new!15} 4$\rightarrow$1  & 75.3 hours  & \textbf{40.6} & \textbf{67.6} & \textbf{77.5} \\
            \midrule
            \textit{Multiple-frame selection} \\
            4$\rightarrow$2  & 77.8 hours  & 41.2 & 67.8 & 77.9\\
            4$\rightarrow$3  & 80.2 hours  & 41.5 & 68.0 & 78.0 \\
            4  & 82.5 hours & 41.7 & 68.3 & 78.3  \\
            8$\rightarrow$4  & 100.3 hours & \textbf{42.8} & \textbf{69.2} & \textbf{79.5} 
        \end{tabular}
}
        \caption{Ablation study on selecting different frames by the proposed selector $\mathcal{S}$. Note that ``4$\rightarrow$1" means selecting one frame among 4 input frames.}
          \label{table_a3}
          \vspace{-3mm}
\end{table}

\paragraph{Impact of masking ratios.} In order to figure out the influence of masking ratios of masked autoencoder for video-language model pre-training, we show the results in Table \ref{table_a1}, when the masking ratio is 50\%, the pre-training time can be efficiently reduced from \textbf{70.8} hours, whose masking ratio is 10\%, to \textbf{50.4} hours, while losing only \textbf{1.2\%} on R@1, which can demonstrate the potential of extending masked autoencoder for video-language pre-training. 

\paragraph{Effect of keeping rates.} The results of adopting different keeping rates for visual token sparsification are shown in Table \ref{table_a2}. When the keeping rate is 0.8 (\ie, 80\%), the pre-training time can be reduced from \textbf{53.8} hours, whose keeping rate is 0.9, to \textbf{50.4} hours, while only losing \textbf{0.5\%} on R@1. The saving of pre-training time is promising, considering the masked autoencoder already removes plenty of  visual patches and the number of the input frames is only 1.

\paragraph{Effectiveness of frame selection.} To evaluate the effectiveness of the proposed selector $\mathcal{S}$ for frame selection, we display the results in Table \ref{table_a3}. When selecting one frame among 4 video frames (\textit{i.e.}, 4$\rightarrow$1), the performance can be improved from \textbf{39.3\%} to \textbf{40.6\%} on R@1. When the selected frames are multiple, for example, when selecting 2 frames among 4 video frames (\textit{i.e.}, 4$\rightarrow$2), the loss on R@1 is only \textbf{0.5\%} compared with the models pre-trained with 4 video frames, while reducing the pre-training time from \textbf{82.5} hours to \textbf{77.8} hours. 

\paragraph{Impact of different components.} In Table \ref{table_a4}, we show the results of adopting different components to figure out their influences. The final pre-training time can be consistently reduced from \textbf{144.5} hours to \textbf{77.8} hours (\textbf{1.9$\times$} speedup), while only losing \textbf{1.4\%} on R@1, which can show the efficiencies of the proposed components.

\begin{figure}[t]
  \hspace{-1mm}
\centering
\includegraphics[width=1\linewidth]{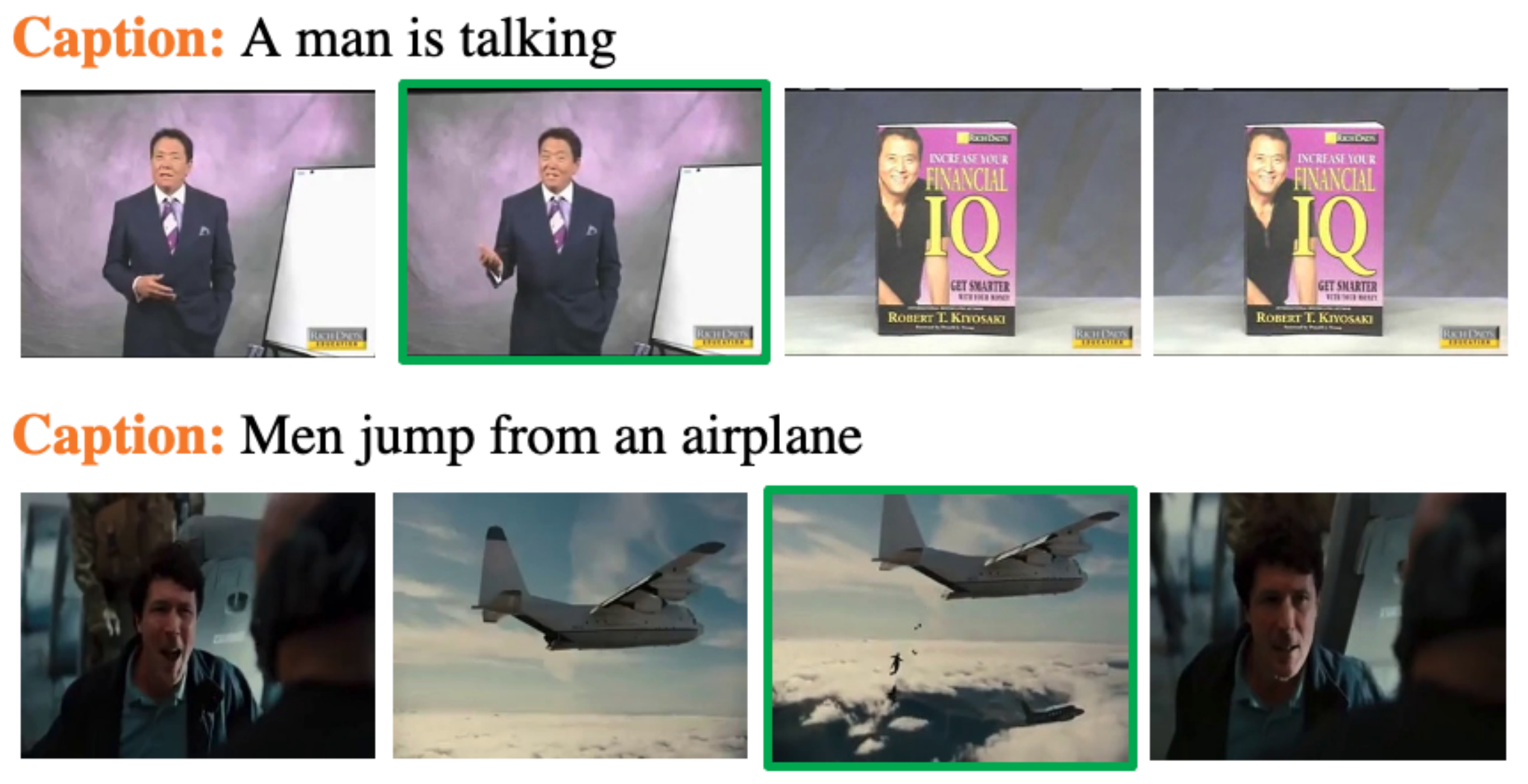}
\vspace{-6.5mm}
    \caption{\textbf{The examples of frame selection.} We show two examples to demonstrate the effectiveness of our frame selector $\mathcal{S}$. Note that in each row, the frame denoted by the \textit{green} box means the final selected frame among the given clip. }
\label{fig:frame_select}
 \vspace{-3mm}
\end{figure}

\begin{figure}[t]
  \hspace{-1mm}
\centering
\includegraphics[width=1\linewidth]{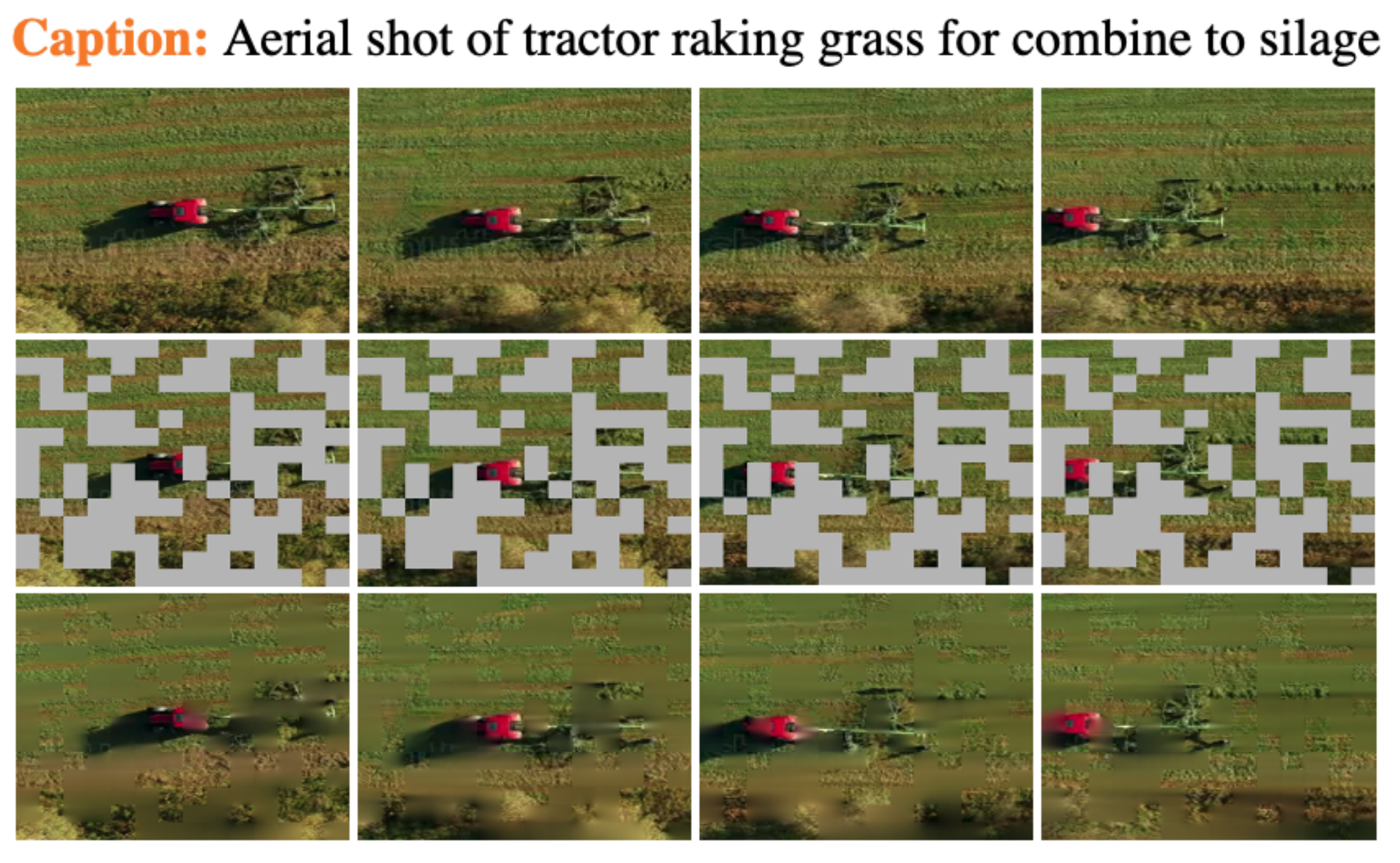}
\vspace{-6.5mm}
    \caption{\textbf{The examples of pixel prediction for masked patches.} The top row represents the original frames, the middle row means the masked frames, and the bottom row denotes the pixel reconstruction for masked patches.}
\label{fig:frame_reconstruct}
 \vspace{-3mm}
\end{figure}

\begin{table}[t]
\resizebox{0.47\textwidth}{!}{%
\centering
      \centering
        \begin{tabular}{cccp{2cm}<{\centering}ccc}
        \multicolumn{1}{c}{\multirow{2}{*}{MAE}} & \multicolumn{1}{c}{\multirow{2}{*}{VTS}} & \multicolumn{1}{c}{\multirow{2}{*}{FS}} & \multicolumn{1}{c}{\multirow{2}{*}{PT Time}}  &\multicolumn{3}{c}{MSRVTT} \\ 
        \multicolumn{1}{c}{} & \multicolumn{1}{c}{} & \multicolumn{1}{c}{} & \multicolumn{1}{c}{} & R@1 & R@5 & R@10 \\  
            \midrule
            \xmark & \xmark & \xmark & 144.5 hours & \textbf{42.6} & \textbf{68.8} &  \textbf{79.2}  \\
            \cmark & \xmark & \xmark & 93.5 hours  & 42.0 & 68.4 &  78.7   \\
            \cmark & \cmark &  \xmark & 82.5 hours & 41.6 & 67.9 & 78.3  \\
            \cmark & \cmark & \cmark & \textbf{77.8 hours} & 41.2  & 67.8  & 77.9 
        \end{tabular}
}
        \caption{Ablation study on using different components of the proposed SMAUG on multiple-frame setting (\textit{i.e.}, ``4$\rightarrow$2"). ``MAE" means whether using masked autoencoder or not, ``VTS" means visual token sparsification, ``FS" denotes frame selection.}
          \label{table_a4}
                    \vspace{-1mm}
\end{table}

\begin{table}[t]
\resizebox{0.48\textwidth}{!}{%
\centering
      \centering
        \begin{tabular}{p{3cm}<{\centering}cccc}
        \multicolumn{1}{c}{\multirow{2}{*}{Method}} & \multicolumn{1}{c}{\multirow{2}{*}{PT Time}} &\multicolumn{3}{c}{MSRVTT} \\ 
        \multicolumn{1}{c}{} &  & R@1 & R@5 & R@10 \\  
            \midrule
            Singularity$^{*}$ & 83.4 hours & 36.8 & 65.9 & 75.5 \\
            Ours$^{*}$ & \textbf{75.3 hours} & \textbf{40.6} & \textbf{67.6} & \textbf{77.5} \\
            \hline
            Singularity$^{+}$ & 285.3 hours & 41.5  &  68.7  & 77.0  \\
            Ours$^{+}$ & \textbf{198.2 hours} & \textbf{44.0} & \textbf{70.4} & \textbf{78.8}\\
        \end{tabular}
}
        \caption{Pre-training with different scales of the video-text corpus. ``*" and ``+" mean the pre-trained models with 5M and 17M corpus respectively. We report the results pre-trained with the single-frame setting, \ie, the number of the final video frames is 1.}
          \label{table_a5}
          \vspace{-5mm}
\end{table}

\subsection{Analysis of SMAUG}
\label{analysis}
\paragraph{Pre-training Datasets and Costs.} To evaluate the pre-training efficiency of our proposed method, we show the results in Table \ref{table_a5}. We report the pre-training time of Singularity \cite{lei2022revealing} by using their official codes. When adopting different dataset corpus (\textit{e.g.}, 5M and 17M) for pre-training, our approach can consistently use less pre-training time while achieving better performances. For example, when pre-trained on the 17M corpus, our approach can obtain \textbf{2.5\%} performance gain on R@1 while saving \textbf{87.1} pre-training hours compared with Singularity. These results can demonstrate the strong scalability of our method.

\paragraph{Visualization.} Finally, we showcase the examples of frame selection by selector $\mathcal{S}$ in Figure \ref{fig:frame_select}. It can be observed that there're abundant video frames which not correspond to the given caption, removing these chaotic frames are reasonable. Furthermore, our approach can accurately select the most essential frames according to the given captions, which can strongly demonstrate its efficiency. 

In Figure \ref{fig:frame_reconstruct}, we display the qualitative examples of pixel reconstruction for masked patches. Even with a masking ratio of 50\%, the model can still produce reliable predictions for masked patches, which can further demonstrate  mask autoencoders can help us achieve satisfactory performances and efficiently mitigate the computational burden of video-language model pre-training.

\section{Conclusion}
In this paper, we propose an efficient video-language pre-training method called SMAUG, which is built on masked autoencoders. In addition, we further propose a space-time sparsification module to remove the spatial and temporal redundancies among the remaining visible patches. Our SMAUG can achieve state-of-the-art or comparable performances on two video-language tasks across six popular benchmarks, while obtaining 1.9$\times$ pre-training time speedup.

{\small
\bibliographystyle{ieee_fullname}
\bibliography{smaug}

\begin{thebibliography}{10}\itemsep=-1pt

\bibitem{anne2017localizing}
Lisa Anne~Hendricks, Oliver Wang, Eli Shechtman, Josef Sivic, Trevor Darrell,
  and Bryan Russell.
\newblock Localizing moments in video with natural language.
\newblock In {\em Proceedings of the IEEE international conference on computer
  vision}, pages 5803--5812, 2017.

\bibitem{bain2021frozen}
Max Bain, Arsha Nagrani, G{\"u}l Varol, and Andrew Zisserman.
\newblock Frozen in time: A joint video and image encoder for end-to-end
  retrieval.
\newblock In {\em Proceedings of the IEEE/CVF International Conference on
  Computer Vision}, pages 1728--1738, 2021.

\bibitem{bao2021beit}
Hangbo Bao, Li Dong, and Furu Wei.
\newblock Beit: Bert pre-training of image transformers.
\newblock {\em arXiv preprint arXiv:2106.08254}, 2021.

\bibitem{buch2022revisiting}
Shyamal Buch, Crist{\'o}bal Eyzaguirre, Adrien Gaidon, Jiajun Wu, Li Fei-Fei,
  and Juan~Carlos Niebles.
\newblock Revisiting the" video" in video-language understanding.
\newblock In {\em Proceedings of the IEEE/CVF Conference on Computer Vision and
  Pattern Recognition}, pages 2917--2927, 2022.

\bibitem{caba2015activitynet}
Fabian Caba~Heilbron, Victor Escorcia, Bernard Ghanem, and Juan Carlos~Niebles.
\newblock Activitynet: A large-scale video benchmark for human activity
  understanding.
\newblock In {\em Proceedings of the ieee conference on computer vision and
  pattern recognition}, pages 961--970, 2015.

\bibitem{changpinyo2021conceptual}
Soravit Changpinyo, Piyush Sharma, Nan Ding, and Radu Soricut.
\newblock Conceptual 12m: Pushing web-scale image-text pre-training to
  recognize long-tail visual concepts.
\newblock In {\em Proceedings of the IEEE/CVF Conference on Computer Vision and
  Pattern Recognition}, pages 3558--3568, 2021.

\bibitem{chen2020generative}
Mark Chen, Alec Radford, Rewon Child, Jeffrey Wu, Heewoo Jun, David Luan, and
  Ilya Sutskever.
\newblock Generative pretraining from pixels.
\newblock In {\em International conference on machine learning}, pages
  1691--1703. PMLR, 2020.

\bibitem{dosovitskiy2020image}
Alexey Dosovitskiy, Lucas Beyer, Alexander Kolesnikov, Dirk Weissenborn,
  Xiaohua Zhai, Thomas Unterthiner, Mostafa Dehghani, Matthias Minderer, Georg
  Heigold, Sylvain Gelly, et~al.
\newblock An image is worth 16x16 words: Transformers for image recognition at
  scale.
\newblock {\em arXiv preprint arXiv:2010.11929}, 2020.

\bibitem{dou2022empirical}
Zi-Yi Dou, Yichong Xu, Zhe Gan, Jianfeng Wang, Shuohang Wang, Lijuan Wang,
  Chenguang Zhu, Pengchuan Zhang, Lu Yuan, Nanyun Peng, et~al.
\newblock An empirical study of training end-to-end vision-and-language
  transformers.
\newblock In {\em Proceedings of the IEEE/CVF Conference on Computer Vision and
  Pattern Recognition}, pages 18166--18176, 2022.

\bibitem{dzabraev2021mdmmt}
Maksim Dzabraev, Maksim Kalashnikov, Stepan Komkov, and Aleksandr Petiushko.
\newblock Mdmmt: Multidomain multimodal transformer for video retrieval.
\newblock In {\em Proceedings of the IEEE/CVF Conference on Computer Vision and
  Pattern Recognition}, pages 3354--3363, 2021.

\bibitem{feichtenhofer2022masked}
Christoph Feichtenhofer, Haoqi Fan, Yanghao Li, and Kaiming He.
\newblock Masked autoencoders as spatiotemporal learners.
\newblock {\em arXiv preprint arXiv:2205.09113}, 2022.

\bibitem{fu2021violet}
Tsu-Jui Fu, Linjie Li, Zhe Gan, Kevin Lin, William~Yang Wang, Lijuan Wang, and
  Zicheng Liu.
\newblock Violet: End-to-end video-language transformers with masked
  visual-token modeling.
\newblock {\em arXiv preprint arXiv:2111.12681}, 2021.

\bibitem{fu2022empirical}
Tsu-Jui Fu, Linjie Li, Zhe Gan, Kevin Lin, William~Yang Wang, Lijuan Wang, and
  Zicheng Liu.
\newblock An empirical study of end-to-end video-language transformers with
  masked visual modeling.
\newblock {\em arXiv preprint arXiv:2209.01540}, 2022.

\bibitem{gabeur2020multi}
Valentin Gabeur, Chen Sun, Karteek Alahari, and Cordelia Schmid.
\newblock Multi-modal transformer for video retrieval.
\newblock In {\em European Conference on Computer Vision}, pages 214--229.
  Springer, 2020.

\bibitem{gao2020listen}
Ruohan Gao, Tae-Hyun Oh, Kristen Grauman, and Lorenzo Torresani.
\newblock Listen to look: Action recognition by previewing audio.
\newblock In {\em Proceedings of the IEEE/CVF Conference on Computer Vision and
  Pattern Recognition}, pages 10457--10467, 2020.

\bibitem{ghodrati2021frameexit}
Amir Ghodrati, Babak~Ehteshami Bejnordi, and Amirhossein Habibian.
\newblock Frameexit: Conditional early exiting for efficient video recognition.
\newblock In {\em Proceedings of the IEEE/CVF Conference on Computer Vision and
  Pattern Recognition}, pages 15608--15618, 2021.

\bibitem{he2022masked}
Kaiming He, Xinlei Chen, Saining Xie, Yanghao Li, Piotr Doll{\'a}r, and Ross
  Girshick.
\newblock Masked autoencoders are scalable vision learners.
\newblock In {\em Proceedings of the IEEE/CVF Conference on Computer Vision and
  Pattern Recognition}, pages 16000--16009, 2022.

\bibitem{jang2016categorical}
Eric Jang, Shixiang Gu, and Ben Poole.
\newblock Categorical reparameterization with gumbel-softmax.
\newblock {\em arXiv preprint arXiv:1611.01144}, 2016.

\bibitem{jiang2022pseudo}
Haojun Jiang, Yuanze Lin, Dongchen Han, Shiji Song, and Gao Huang.
\newblock Pseudo-q: Generating pseudo language queries for visual grounding.
\newblock In {\em Proceedings of the IEEE/CVF Conference on Computer Vision and
  Pattern Recognition}, pages 15513--15523, 2022.

\bibitem{kim2021efficient}
Hanul Kim, Mihir Jain, Jun-Tae Lee, Sungrack Yun, and Fatih Porikli.
\newblock Efficient action recognition via dynamic knowledge propagation.
\newblock In {\em Proceedings of the IEEE/CVF International Conference on
  Computer Vision}, pages 13719--13728, 2021.

\bibitem{korbar2019scsampler}
Bruno Korbar, Du Tran, and Lorenzo Torresani.
\newblock Scsampler: Sampling salient clips from video for efficient action
  recognition.
\newblock In {\em Proceedings of the IEEE/CVF International Conference on
  Computer Vision}, pages 6232--6242, 2019.

\bibitem{krishna2017dense}
Ranjay Krishna, Kenji Hata, Frederic Ren, Li Fei-Fei, and Juan Carlos~Niebles.
\newblock Dense-captioning events in videos.
\newblock In {\em Proceedings of the IEEE international conference on computer
  vision}, pages 706--715, 2017.

\bibitem{krishna2017visual}
Ranjay Krishna, Yuke Zhu, Oliver Groth, Justin Johnson, Kenji Hata, Joshua
  Kravitz, Stephanie Chen, Yannis Kalantidis, Li-Jia Li, David~A Shamma, et~al.
\newblock Visual genome: Connecting language and vision using crowdsourced
  dense image annotations.
\newblock {\em International journal of computer vision}, 123(1):32--73, 2017.

\bibitem{kwon2022masked}
Gukyeong Kwon, Zhaowei Cai, Avinash Ravichandran, Erhan Bas, Rahul Bhotika, and
  Stefano Soatto.
\newblock Masked vision and language modeling for multi-modal representation
  learning.
\newblock {\em arXiv preprint arXiv:2208.02131}, 2022.

\bibitem{le2020hierarchical}
Thao~Minh Le, Vuong Le, Svetha Venkatesh, and Truyen Tran.
\newblock Hierarchical conditional relation networks for video question
  answering.
\newblock In {\em Proceedings of the IEEE/CVF conference on computer vision and
  pattern recognition}, pages 9972--9981, 2020.

\bibitem{lei2022revealing}
Jie Lei, Tamara~L Berg, and Mohit Bansal.
\newblock Revealing single frame bias for video-and-language learning.
\newblock {\em arXiv preprint arXiv:2206.03428}, 2022.

\bibitem{lei2021less}
Jie Lei, Linjie Li, Luowei Zhou, Zhe Gan, Tamara~L Berg, Mohit Bansal, and
  Jingjing Liu.
\newblock Less is more: Clipbert for video-and-language learning via sparse
  sampling.
\newblock In {\em Proceedings of the IEEE/CVF Conference on Computer Vision and
  Pattern Recognition}, pages 7331--7341, 2021.

\bibitem{li2022align}
Dongxu Li, Junnan Li, Hongdong Li, Juan~Carlos Niebles, and Steven~CH Hoi.
\newblock Align and prompt: Video-and-language pre-training with entity
  prompts.
\newblock In {\em Proceedings of the IEEE/CVF Conference on Computer Vision and
  Pattern Recognition}, pages 4953--4963, 2022.

\bibitem{li2020hero}
Linjie Li, Yen-Chun Chen, Yu Cheng, Zhe Gan, Licheng Yu, and Jingjing Liu.
\newblock Hero: Hierarchical encoder for video+ language omni-representation
  pre-training.
\newblock {\em arXiv preprint arXiv:2005.00200}, 2020.

\bibitem{liang2022not}
Youwei Liang, Chongjian Ge, Zhan Tong, Yibing Song, Jue Wang, and Pengtao Xie.
\newblock Not all patches are what you need: Expediting vision transformers via
  token reorganizations.
\newblock {\em arXiv preprint arXiv:2202.07800}, 2022.

\bibitem{lin2014microsoft}
Tsung-Yi Lin, Michael Maire, Serge Belongie, James Hays, Pietro Perona, Deva
  Ramanan, Piotr Doll{\'a}r, and C~Lawrence Zitnick.
\newblock Microsoft coco: Common objects in context.
\newblock In {\em European conference on computer vision}, pages 740--755.
  Springer, 2014.

\bibitem{lin2021self}
Yuanze Lin, Xun Guo, and Yan Lu.
\newblock Self-supervised video representation learning with meta-contrastive
  network.
\newblock In {\em Proceedings of the IEEE/CVF International Conference on
  Computer Vision}, pages 8239--8249, 2021.

\bibitem{lin2022revive}
Yuanze Lin, Yujia Xie, Dongdong Chen, Yichong Xu, Chenguang Zhu, and Lu Yuan.
\newblock Revive: Regional visual representation matters in knowledge-based
  visual question answering.
\newblock {\em arXiv preprint arXiv:2206.01201}, 2022.

\bibitem{loshchilov2016sgdr}
Ilya Loshchilov and Frank Hutter.
\newblock Sgdr: Stochastic gradient descent with warm restarts.
\newblock {\em arXiv preprint arXiv:1608.03983}, 2016.

\bibitem{loshchilov2017decoupled}
Ilya Loshchilov and Frank Hutter.
\newblock Decoupled weight decay regularization.
\newblock {\em arXiv preprint arXiv:1711.05101}, 2017.

\bibitem{luo2020univl}
Huaishao Luo, Lei Ji, Botian Shi, Haoyang Huang, Nan Duan, Tianrui Li, Jason
  Li, Taroon Bharti, and Ming Zhou.
\newblock Univl: A unified video and language pre-training model for multimodal
  understanding and generation.
\newblock {\em arXiv preprint arXiv:2002.06353}, 2020.

\bibitem{luo2021clip4clip}
Huaishao Luo, Lei Ji, Ming Zhong, Yang Chen, Wen Lei, Nan Duan, and Tianrui Li.
\newblock Clip4clip: An empirical study of clip for end to end video clip
  retrieval.
\newblock {\em arXiv preprint arXiv:2104.08860}, 2021.

\bibitem{meng2020ar}
Yue Meng, Chung-Ching Lin, Rameswar Panda, Prasanna Sattigeri, Leonid
  Karlinsky, Aude Oliva, Kate Saenko, and Rogerio Feris.
\newblock Ar-net: Adaptive frame resolution for efficient action recognition.
\newblock In {\em European Conference on Computer Vision}, pages 86--104.
  Springer, 2020.

\bibitem{miech2019howto100m}
Antoine Miech, Dimitri Zhukov, Jean-Baptiste Alayrac, Makarand Tapaswi, Ivan
  Laptev, and Josef Sivic.
\newblock Howto100m: Learning a text-video embedding by watching hundred
  million narrated video clips.
\newblock In {\em Proceedings of the IEEE/CVF International Conference on
  Computer Vision}, pages 2630--2640, 2019.

\bibitem{ordonez2011im2text}
Vicente Ordonez, Girish Kulkarni, and Tamara Berg.
\newblock Im2text: Describing images using 1 million captioned photographs.
\newblock {\em Advances in neural information processing systems}, 24, 2011.

\bibitem{patrick2020support}
Mandela Patrick, Po-Yao Huang, Yuki Asano, Florian Metze, Alexander Hauptmann,
  Joao Henriques, and Andrea Vedaldi.
\newblock Support-set bottlenecks for video-text representation learning.
\newblock {\em arXiv preprint arXiv:2010.02824}, 2020.

\bibitem{radford2021learning}
Alec Radford, Jong~Wook Kim, Chris Hallacy, Aditya Ramesh, Gabriel Goh,
  Sandhini Agarwal, Girish Sastry, Amanda Askell, Pamela Mishkin, Jack Clark,
  et~al.
\newblock Learning transferable visual models from natural language
  supervision.
\newblock In {\em International Conference on Machine Learning}, pages
  8748--8763. PMLR, 2021.

\bibitem{rao2021dynamicvit}
Yongming Rao, Wenliang Zhao, Benlin Liu, Jiwen Lu, Jie Zhou, and Cho-Jui Hsieh.
\newblock Dynamicvit: Efficient vision transformers with dynamic token
  sparsification.
\newblock {\em Advances in neural information processing systems},
  34:13937--13949, 2021.

\bibitem{rouditchenko2020avlnet}
Andrew Rouditchenko, Angie Boggust, David Harwath, Brian Chen, Dhiraj Joshi,
  Samuel Thomas, Kartik Audhkhasi, Hilde Kuehne, Rameswar Panda, Rogerio Feris,
  et~al.
\newblock Avlnet: Learning audio-visual language representations from
  instructional videos.
\newblock {\em arXiv preprint arXiv:2006.09199}, 2020.

\bibitem{sennrich2015neural}
Rico Sennrich, Barry Haddow, and Alexandra Birch.
\newblock Neural machine translation of rare words with subword units.
\newblock {\em arXiv preprint arXiv:1508.07909}, 2015.

\bibitem{sharma2018conceptual}
Piyush Sharma, Nan Ding, Sebastian Goodman, and Radu Soricut.
\newblock Conceptual captions: A cleaned, hypernymed, image alt-text dataset
  for automatic image captioning.
\newblock In {\em Proceedings of the 56th Annual Meeting of the Association for
  Computational Linguistics (Volume 1: Long Papers)}, pages 2556--2565, 2018.

\bibitem{suhr2018corpus}
Alane Suhr, Stephanie Zhou, Ally Zhang, Iris Zhang, Huajun Bai, and Yoav Artzi.
\newblock A corpus for reasoning about natural language grounded in
  photographs.
\newblock {\em arXiv preprint arXiv:1811.00491}, 2018.

\bibitem{sun2021dynamic}
Ximeng Sun, Rameswar Panda, Chun-Fu~Richard Chen, Aude Oliva, Rogerio Feris,
  and Kate Saenko.
\newblock Dynamic network quantization for efficient video inference.
\newblock In {\em Proceedings of the IEEE/CVF International Conference on
  Computer Vision}, pages 7375--7385, 2021.

\bibitem{tong2022videomae}
Zhan Tong, Yibing Song, Jue Wang, and Limin Wang.
\newblock Videomae: Masked autoencoders are data-efficient learners for
  self-supervised video pre-training.
\newblock {\em arXiv preprint arXiv:2203.12602}, 2022.

\bibitem{vaswani2017attention}
Ashish Vaswani, Noam Shazeer, Niki Parmar, Jakob Uszkoreit, Llion Jones,
  Aidan~N Gomez, {\L}ukasz Kaiser, and Illia Polosukhin.
\newblock Attention is all you need.
\newblock {\em Advances in neural information processing systems}, 30, 2017.

\bibitem{vincent2008extracting}
Pascal Vincent, Hugo Larochelle, Yoshua Bengio, and Pierre-Antoine Manzagol.
\newblock Extracting and composing robust features with denoising autoencoders.
\newblock In {\em Proceedings of the 25th international conference on Machine
  learning}, pages 1096--1103, 2008.

\bibitem{vincent2010stacked}
Pascal Vincent, Hugo Larochelle, Isabelle Lajoie, Yoshua Bengio, Pierre-Antoine
  Manzagol, and L{\'e}on Bottou.
\newblock Stacked denoising autoencoders: Learning useful representations in a
  deep network with a local denoising criterion.
\newblock {\em Journal of machine learning research}, 11(12), 2010.

\bibitem{wang2022all}
Alex~Jinpeng Wang, Yixiao Ge, Rui Yan, Yuying Ge, Xudong Lin, Guanyu Cai,
  Jianping Wu, Ying Shan, Xiaohu Qie, and Mike~Zheng Shou.
\newblock All in one: Exploring unified video-language pre-training.
\newblock {\em arXiv preprint arXiv:2203.07303}, 2022.

\bibitem{wang2018reconstruction}
Bairui Wang, Lin Ma, Wei Zhang, and Wei Liu.
\newblock Reconstruction network for video captioning.
\newblock In {\em Proceedings of the IEEE conference on computer vision and
  pattern recognition}, pages 7622--7631, 2018.

\bibitem{wang2022bevt}
Rui Wang, Dongdong Chen, Zuxuan Wu, Yinpeng Chen, Xiyang Dai, Mengchen Liu,
  Yu-Gang Jiang, Luowei Zhou, and Lu Yuan.
\newblock Bevt: Bert pretraining of video transformers.
\newblock In {\em Proceedings of the IEEE/CVF Conference on Computer Vision and
  Pattern Recognition}, pages 14733--14743, 2022.

\bibitem{wang2019vatex}
Xin Wang, Jiawei Wu, Junkun Chen, Lei Li, Yuan-Fang Wang, and William~Yang
  Wang.
\newblock Vatex: A large-scale, high-quality multilingual dataset for
  video-and-language research.
\newblock In {\em Proceedings of the IEEE/CVF International Conference on
  Computer Vision}, pages 4581--4591, 2019.

\bibitem{wang2022adafocus}
Yulin Wang, Yang Yue, Yuanze Lin, Haojun Jiang, Zihang Lai, Victor Kulikov,
  Nikita Orlov, Humphrey Shi, and Gao Huang.
\newblock Adafocus v2: End-to-end training of spatial dynamic networks for
  video recognition.
\newblock In {\em 2022 IEEE/CVF Conference on Computer Vision and Pattern
  Recognition (CVPR)}, pages 20030--20040. IEEE, 2022.

\bibitem{wei2022masked}
Chen Wei, Haoqi Fan, Saining Xie, Chao-Yuan Wu, Alan Yuille, and Christoph
  Feichtenhofer.
\newblock Masked feature prediction for self-supervised visual pre-training.
\newblock In {\em Proceedings of the IEEE/CVF Conference on Computer Vision and
  Pattern Recognition}, pages 14668--14678, 2022.

\bibitem{xu2017video}
Dejing Xu, Zhou Zhao, Jun Xiao, Fei Wu, Hanwang Zhang, Xiangnan He, and Yueting
  Zhuang.
\newblock Video question answering via gradually refined attention over
  appearance and motion.
\newblock In {\em Proceedings of the 25th ACM international conference on
  Multimedia}, pages 1645--1653, 2017.

\bibitem{xu2021videoclip}
Hu Xu, Gargi Ghosh, Po-Yao Huang, Dmytro Okhonko, Armen Aghajanyan, Florian
  Metze, Luke Zettlemoyer, and Christoph Feichtenhofer.
\newblock Videoclip: Contrastive pre-training for zero-shot video-text
  understanding.
\newblock {\em arXiv preprint arXiv:2109.14084}, 2021.

\bibitem{xu2016msr}
Jun Xu, Tao Mei, Ting Yao, and Yong Rui.
\newblock Msr-vtt: A large video description dataset for bridging video and
  language.
\newblock In {\em Proceedings of the IEEE conference on computer vision and
  pattern recognition}, pages 5288--5296, 2016.

\bibitem{yang2021just}
Antoine Yang, Antoine Miech, Josef Sivic, Ivan Laptev, and Cordelia Schmid.
\newblock Just ask: Learning to answer questions from millions of narrated
  videos.
\newblock In {\em Proceedings of the IEEE/CVF International Conference on
  Computer Vision}, pages 1686--1697, 2021.

\bibitem{yu2018joint}
Youngjae Yu, Jongseok Kim, and Gunhee Kim.
\newblock A joint sequence fusion model for video question answering and
  retrieval.
\newblock In {\em Proceedings of the European Conference on Computer Vision
  (ECCV)}, pages 471--487, 2018.

\bibitem{yu2019activitynet}
Zhou Yu, Dejing Xu, Jun Yu, Ting Yu, Zhou Zhao, Yueting Zhuang, and Dacheng
  Tao.
\newblock Activitynet-qa: A dataset for understanding complex web videos via
  question answering.
\newblock In {\em Proceedings of the AAAI Conference on Artificial
  Intelligence}, volume~33, pages 9127--9134, 2019.

\bibitem{zellers2021merlot}
Rowan Zellers, Ximing Lu, Jack Hessel, Youngjae Yu, Jae~Sung Park, Jize Cao,
  Ali Farhadi, and Yejin Choi.
\newblock Merlot: Multimodal neural script knowledge models.
\newblock {\em Advances in Neural Information Processing Systems},
  34:23634--23651, 2021.

\bibitem{zhang2021vinvl}
Pengchuan Zhang, Xiujun Li, Xiaowei Hu, Jianwei Yang, Lei Zhang, Lijuan Wang,
  Yejin Choi, and Jianfeng Gao.
\newblock Vinvl: Revisiting visual representations in vision-language models.
\newblock In {\em Proceedings of the IEEE/CVF Conference on Computer Vision and
  Pattern Recognition}, pages 5579--5588, 2021.

\bibitem{zhou2018towards}
Luowei Zhou, Chenliang Xu, and Jason~J Corso.
\newblock Towards automatic learning of procedures from web instructional
  videos.
\newblock In {\em Thirty-Second AAAI Conference on Artificial Intelligence},
  2018.

\bibitem{zhu2020actbert}
Linchao Zhu and Yi Yang.
\newblock Actbert: Learning global-local video-text representations.
\newblock In {\em Proceedings of the IEEE/CVF conference on computer vision and
  pattern recognition}, pages 8746--8755, 2020.

\end{thebibliography}
}

\newpage
\appendix 

\section{Overview}
In this supplementary material, we provide the following sections:

   - Details of visual token sparsification in Section \ref{sparsification}.

   - Details of pre-training objectives in Section \ref{objectives2}.
   
   - Additional ablations of SMAUG in Section \ref{ablation}.
   
   - Comprehensive visualization results in Section \ref{visualization}.

\section{Visual Token Sparsification}
\label{sparsification}
In this section, we provide the details of the visual token sparsification module, which is motivated by EVIT \cite{liang2022not}. For ViT's \cite{dosovitskiy2020image} attention mechanism \cite{vaswani2017attention}, the interactions between \texttt{[CLS]} token and other tokens of each head in transformer block can be denoted as: 

\begin{equation}
    \label{a1}
    X_{\text{cls}} = \text{Softmax}(\frac{Q_{\text{cls}}K^{T}}{\sqrt{d}})V = a\cdot V, \\
\end{equation}
in which  $Q_{\text{cls}}$ represent \textit{query} vector of \texttt{[CLS]} token, $K$ and $V$ mean \textit{key} and \textit{value} matrices, $d$ is the dimension of query matrice, and $X_{\text{cls}}$ is the output of the \texttt{[CLS]} token. The attention values of all heads are $\{a_{i}\}_{i=1}^{h}$, $h$ means the number of heads, the average attention value of all heads is $\bar{a}=\sum\nolimits_{i=1}^{h} a_{i}/h$. 

We keep the tokens that have top-$k$ elements in $\bar{a}$, these tokens are called attentive tokens, the other tokens are inattentive tokens, the keeping rate is defined as $\gamma=k/p$, where $k$ and $p$ mean the number of attentive tokens and total input tokens, we can adjust the value of $\gamma$ to determine the number of final preserved tokens. 

\paragraph{Inattentive token fusion.} We adopt the attention values to fuse the inattentive tokens, the fusion process can be regarded as a weight average operation: 

\begin{equation}
    \label{a1}
    X_{\text{fused}}=\sum\nolimits_{i\in \mathcal{U}} a_{i}X_{i}, \\
\end{equation}
in which $X_{\text{fused}}$ means fused token, it's then appended into attentive tokens for final preserved tokens. $X_{i}$ and $a_{i}$ represent the inattentive token and attention values from \texttt{[CLS]} token, $\mathcal{U}$ denotes the set of indices of inattentive tokens.

\begin{table*}[t]
{\centering}
\resizebox{1\textwidth}{!}{%
\begin{tabular}{cccccccccccc}
\multicolumn{1}{c}{\multirow{2}{*}{Method}} & \multicolumn{1}{c}{\multirow{2}{*}{PT Dataset}}
& \multicolumn{1}{c}{\multirow{2}{*}{\#Frame}} 
&\multicolumn{3}{c}{MSRVTT} & \multicolumn{3}{c}{DiDeMo} & \multicolumn{3}{c}{ActivityNet Cap} %
\\ 
\multicolumn{1}{c}{} & \multicolumn{1}{c}{} & \multicolumn{1}{c}{}& R@1 & R@5 & R@10 & R@1 & R@5 & R@10 & R@1 & R@5 & R@10 \\ 
\midrule

VIOLET~\cite{fu2021violet} &  YT180M+5M & 4 & 34.5 & 63.0 & 73.4 & 32.6 & 62.8 & 74.7 & - & - & - \\
ClipBERT~\cite{lei2021less} & COCO + VG  & 16/16/8 & 22.0 & 46.8 & 59.9 & 20.4 & 48.0 & 60.8 & 21.3 & 49.0 & 63.5  \\
Frozen~\cite{bain2021frozen} &  5M & 4 & \color{gray}{31.0} & \color{gray}{59.5} & \color{gray}{70.5} & 31.0 & 59.8 & 72.4 & - & - & - \\
ALPRO~\cite{li2022align} &  5M  & 8 & 33.9 & 60.7 & 73.2 & 35.9 & 67.5 & 78.8 & - & - & - \\
Singularity~\cite{lei2022revealing} & 5M  & 1 & 36.8 & 65.9 & 75.5 & 47.4 & 75.2 & 84.0 & 43.0 & 70.6 & 81.3 \\
Singularity-temporal~\cite{lei2022revealing} & 5M  & 4 & 39.9 & 67.3 & 76.0 & 49.2 & 77.5 & 85.4 & 45.9 & 73.3 & 83.8 \\
Singularity~\cite{lei2022revealing} & 17M  & 1 & 41.5 & 68.7 & 77.0 & 53.9 & 79.4 & 86.9 & 47.1 & 75.5 & 85.5 \\
Singularity-temporal~\cite{lei2022revealing} & 17M  & 4 & 42.7 & 69.5 & 78.1 & 53.1 & 79.9 & 88.1 & 48.9 & 77.0 & 86.3 \\
\midrule
\textbf{Ours} & \textbf{5M}  & \textbf{1} & \textbf{40.6} & \textbf{67.6} & \textbf{77.5} & \textbf{49.2} & \textbf{76.7} & \textbf{85.6} & \textbf{44.8}  & \textbf{72.2}  &  \textbf{82.7} \\
\textbf{Ours} & \textbf{5M} & \textbf{4} & \textbf{42.8} & \textbf{69.2} & \textbf{79.5}  & \textbf{51.4} & \textbf{79.5} & \textbf{86.8} & \textbf{48.3} & \textbf{75.7} & \textbf{85.5} \\
\textbf{Ours} & \textbf{17M}  & \textbf{1} & \textbf{44.0} & \textbf{70.4} & \textbf{78.8} & \textbf{55.6} & \textbf{80.8}  & \textbf{88.4}  &  \textbf{49.2} & \textbf{76.9}  & \textbf{86.8}  \\
\textbf{Ours} & \textbf{17M}  & \textbf{4} & \textbf{45.1} & \textbf{71.2} & \textbf{79.6} & \textbf{55.8} & \textbf{81.3}  & \textbf{89.5}  &  \textbf{50.7} & \textbf{78.2}  & \textbf{87.6}  \\
\end{tabular}
}
\vspace{-2mm}
\caption{
\textbf{Results compared with existing video-language pre-training methods on text-to-retrieval}. The pre-training (PT) text-video datasets are HowTo100M (HT100M)~\cite{miech2019howto100m}, YT-Temporal-180M (YT180M)\cite{zellers2021merlot}, MS-COCO (COCO)~\cite{lin2014microsoft}, Visual Genome (VG)~\cite{krishna2017visual}, 5M corpus and 17M corpus. For MSRVTT dataset, results using 9K training videos are \textbf{{\color{gray}{greyed out}}}. ``\#Frame" denotes the number of final video frames for video-language model pre-training. For those methods which adopt different input frames for different datasets (\eg, ClipBERT~\cite{lei2021less}), we leverage ``/" to separate them.} 
\label{table_bx}
\vspace{-2mm}
\end{table*}

\section{Pre-training Objectives}
\label{objectives2}
Here, we further present more details of our utilized pre-training objectives, including video-text contrastive (VTC), video-text matching (VTM), masked language modeling (MLM) and masked visual modeling (MVM). These objectives have been demonstrated effective in video-language pre-training \cite{lei2022revealing,lei2021less,fu2022empirical}.

\paragraph{Video-text contrastive. } The video-text contrastive objective encourages the alignment between corresponding visual and textual embeddings. Given the $i$-th video ($v_{i}$) and $j$-th textual sentence ($t_{j}$) as input, the visual and textual embeddings of \texttt{[CLS]} token are $\mathcal{F}_{\text{v\_cls}}^{i}$ and $\mathcal{F}_{\text{l\_cls}}^{j}$ respectively, we adopt two projection heads to map them into a low-dimentional (\textit{e.g.}, 256-d) space. The similarity score of them can be denoted as:

\begin{equation}
    \label{a1}
    s(v_i,t_j) = \mathcal{P}_{v}(\mathcal{F}_{\text{v\_cls}}^{i})\mathcal{P}_{l}(\mathcal{F}_{\text{l\_cls}}^{j}), \\
\end{equation}
where $s(v_{i},t_{j})$ denotes the similarity score, $\mathcal{P}_{v}$ and $\mathcal{P}_{l}$ are two projectors, each of them consists of a single fully connected layer. The video-text contrastive loss $\mathcal{L}_{\text{vtc}}$ contains two symmetric losses, which can then be represented as:

\begin{equation}
    \label{a2}
    \mathcal{L}_{\text{v2t}}=-\frac{1}{B}\sum\nolimits_{i=1}^{B}\text{log}\frac{\text{exp}(s(v_i,t_i)/\tau)}{\sum\nolimits_{j=1}^{B}\text{exp}(s(v_i,t_j)/\tau)},  \\
\end{equation}
\begin{equation}
    \label{a2_2}
    \mathcal{L}_{\text{t2v}}=-\frac{1}{B}\sum\nolimits_{i=1}^{B}\text{log}\frac{\text{exp}(s(t_i,v_i)/\tau)}{\sum\nolimits_{j=1}^{B}\text{exp}(s(t_i,v_j)/\tau)}, \\
\end{equation}
\begin{equation}
    \label{a3}
    \mathcal{L}_{\text{vtc}} =  \frac{1}{2}(\mathcal{L}_{\text{v2t}}+\mathcal{L}_{\text{t2v}}).
\end{equation}

In Equation (\ref{a2})(\ref{a2_2})(\ref{a3}), $B$ means batch size, $\tau$ is the temperature parameter to modulate the values, we set $\tau$ as 0.07 in our experiments.

\paragraph{Video-text matching. } The objective's goal is to align input videos and texts from the same pairs as well, we use the \texttt{[CLS]} token output from the multi-modal video-text encoder to perform classification, which predicts whether the input videos and texts match or not.
 
\paragraph{Masked language modeling. } It aims to reconstruct the masked textual tokens. Specifically, we also resort to visual contexts, \textit{i.e.}, we use the feature from the last layer of the multi-modal video-text encoder to generate masked textual tokens, the adopted masking ratio of textual tokens is 15\%.

\paragraph{Masked visual modeling. } The utilized masked visual modeling (MVM) objective is learned in mask autoencoders (MAE), the reconstruction target of MVM is pixel values, and we adopt the decoder of SMAUG's masked autoencoder to generate pixels of masked patches.

\begin{table}[t]
\hspace{7mm}
\resizebox{0.37\textwidth}{!}{%
\centering
      \centering
        \begin{tabular}{p{2.5cm}<{\centering}ccc}
        \multicolumn{1}{c}{\multirow{2}{*}{Frame Resolution}} &\multicolumn{3}{c}{MSRVTT} \\
        \multicolumn{1}{c}{} & R@1 & R@5 & R@10 \\  
            \midrule
           \rowcolor{blue_new!15}  224$\times$224 & 39.3 & 65.4 & 75.6  \\
            288$\times$288 & 40.8 & 66.5 & 76.5 \\
            336$\times$336 & \textbf{42.5} & \textbf{67.8} & \textbf{77.8}  \\
        \end{tabular}
}
        \caption{Ablation study on using different frame resolutions for fine-tuning models on down-stream video-language tasks.}
          \label{table_b1}
\end{table}

\section{Additional Ablations of SMAUG}
\label{ablation}
Then, additional ablations are performed to figure out the influences of different components for SMAUG, in Table \ref{table_b1} and \ref{table_b2}, we only sample one video frame for the ablations. The results highlighted in \colorbox{blue_new!20}{blue} are used to specify the default settings. 

\paragraph{Impact of frame resolutions. } The results of using different frame resolutions are reported in Table \ref{table_b1}. We can observe that using a larger frame resolution can significantly improve the model's performance in the fine-tuning stage, especially, there's \textbf{3.2\%} gain on R@1 when changing the frame resolution from 224$\times$224 to 336$\times$336.

\begin{table}[t]
\resizebox{0.47\textwidth}{!}{%
\centering
      \centering
        \begin{tabular}{p{2.5cm}<{\centering}p{2cm}<{\centering}ccc}
        \multicolumn{1}{c}{\multirow{2}{*}{Decoder Depth}} & \multicolumn{1}{c}{\multirow{2}{*}{PT Time}} &\multicolumn{3}{c}{MSRVTT} \\ 
        \multicolumn{1}{c}{} & \multicolumn{1}{c}{} & R@1 & R@5 & R@10 \\  
            \midrule
            1 & 46.8 hours & 37.9 & 64.3 & 74.8  \\
            \rowcolor{blue_new!15}  3 & 50.4 hours & 39.3 & 65.4 & 75.6   \\
            6 & 60.2 hours & 39.8 & 65.9 & 76.0 \\
            12 & 79.7 hours & \textbf{40.4} & \textbf{66.4} & \textbf{76.3} \\
        \end{tabular}
}
        \caption{Ablation study on using different decoder depths on down-stream video-language tasks. We report pre-training (PT) time by using GPU hours, and a single A6000 GPU is adopted.}
          \label{table_b2}
\end{table}

\paragraph{Impact of MAE's decoder depth. } In Table \ref{table_b2}, we report the results of using different decoder depths for masked autoencoder (MAE) \cite{he2022masked}. Decoder depth indicates the number of transformer blocks \cite{dosovitskiy2020image} used in the decoder. Although increasing more transformer blocks in masked autoencoder's decoder can improve the model's performance, it can lead to a significant pre-training time burden. In our experiments, we set the decoder depth as 3, since it can achieve a decent trade-off between the performance and pre-training time.

\paragraph{Comparison with existing methods. } In Table \ref{table_bx}, we show the performance comparison of different methods on text-to-video retrieval, especially, we also report our method and Singularity \cite{lei2022revealing} using single-frame and multiple-frame settings. Our approach can achieve significant performances compared with other methods. As for SMAUG, when ``\#Frame" is 4, we select 4 frames among 8 sampled video frames, our approach can obtain \textbf{45.1\%} R@1 on MSRVTT dataset, which is \textbf{2.4\%} performance gain on R@1 compared with Singularity \cite{lei2022revealing}.

\paragraph{Hyper-parameters. } To make the implementation details clearer, we summarize default hyper-parameters of pre-training and fine-tuning in Table \ref{table_be1} and Table \ref{table_be2}.

\section{Visualization}
\label{visualization}
\begin{figure*}[t]
  \hspace{-1mm}
\centering
\includegraphics[width=0.95\linewidth]{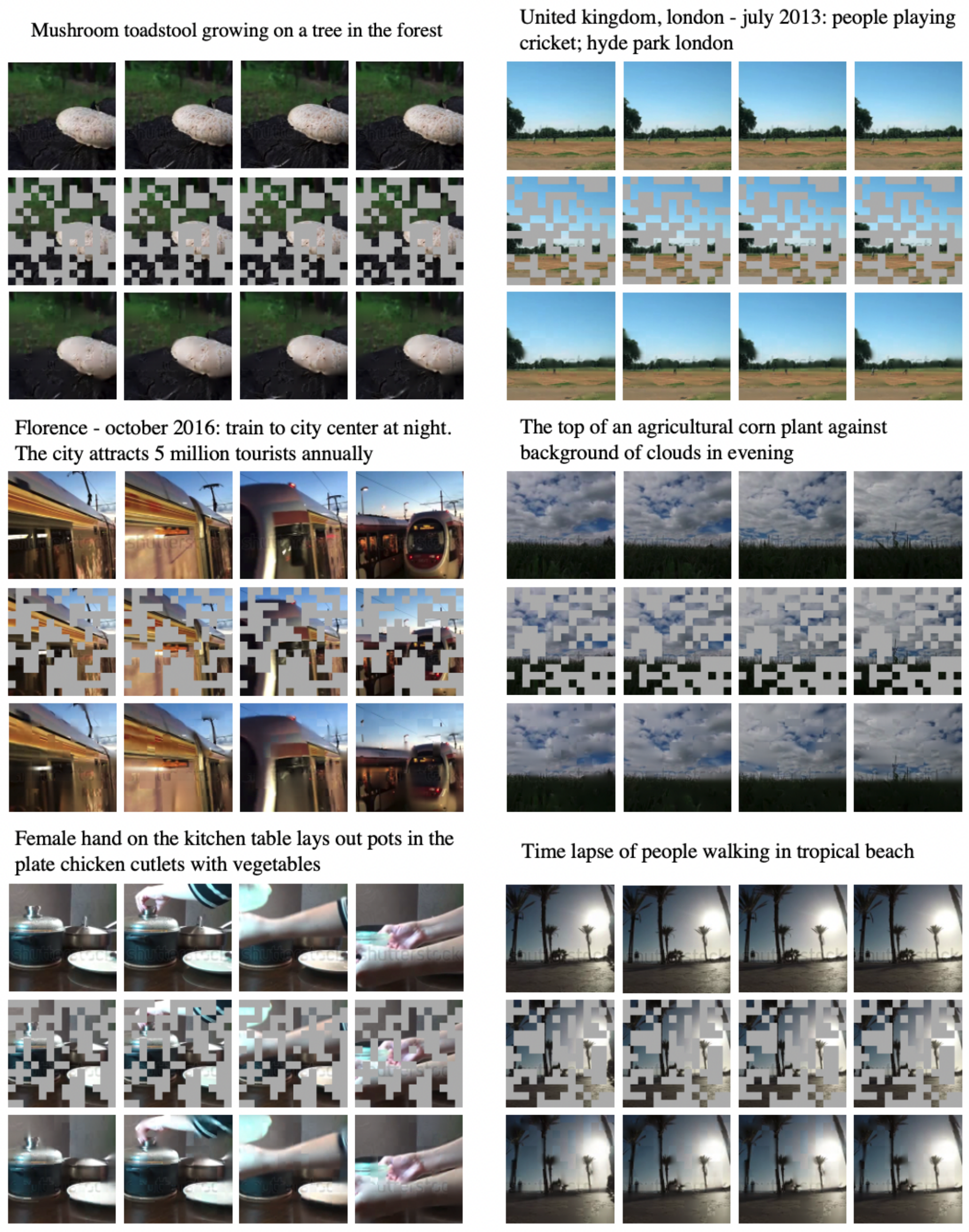}
    \caption{\textbf{The examples of pixel prediction for masked patches.} There're six examples totally, for each sample, the top row represents the original frames, the middle row means the masked frames, and the bottom row denotes the pixel reconstruction for masked patches.}
\label{fig:frame_reconstruct1}
 \vspace{-5mm}
\end{figure*}

\begin{figure*}[t]
  \hspace{-1mm}
\centering
\includegraphics[width=0.95\linewidth]{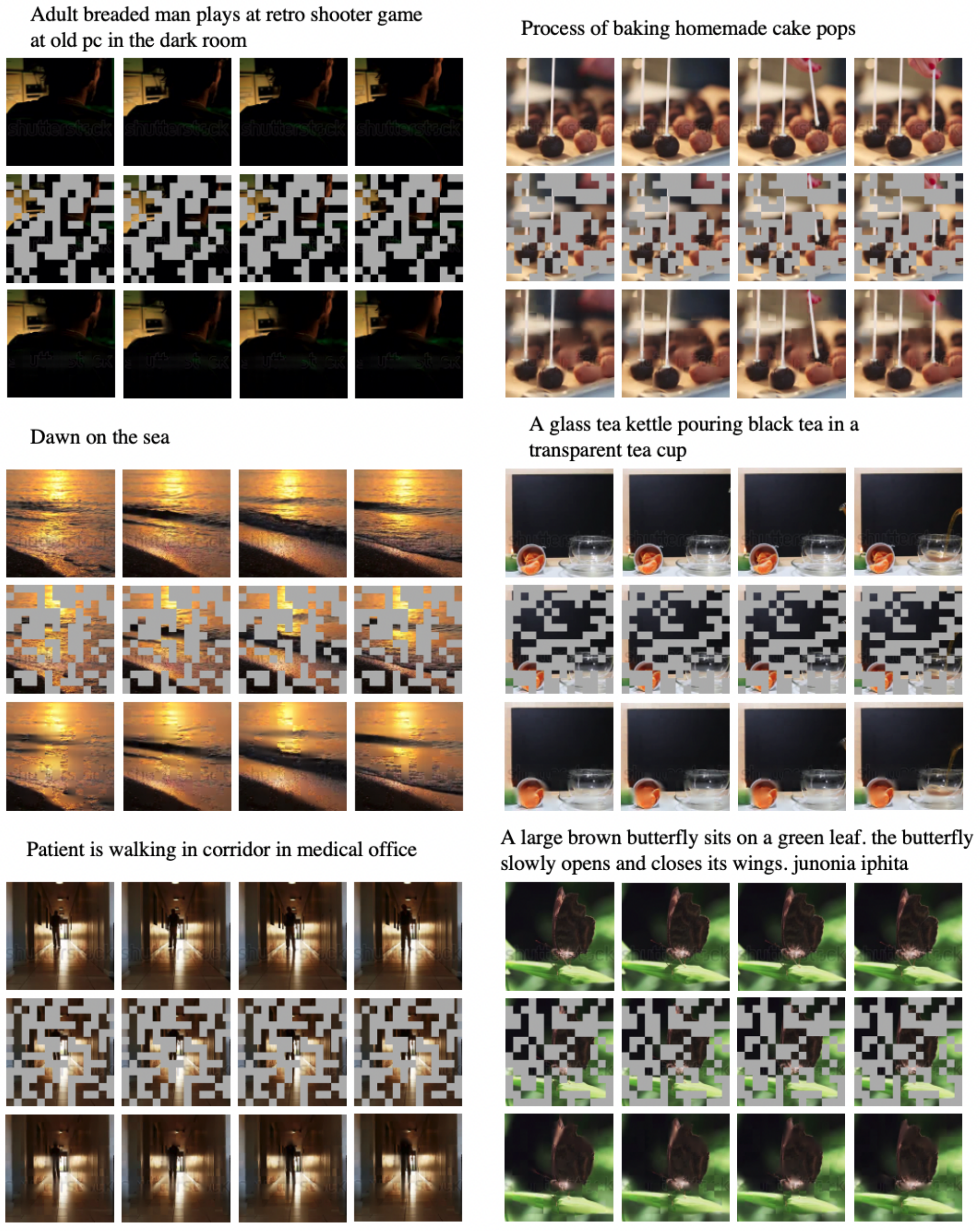}
    \caption{\textbf{The examples of pixel prediction for masked patches.} There're six examples totally, for each sample, the top row represents the original frames, the middle row means the masked frames, and the bottom row denotes the pixel reconstruction for masked patches.}
\label{fig:frame_reconstruct2}
 \vspace{-5mm}
\end{figure*}

Finally, we showcase more visualization results of our SMAUG. Especially, Figure \ref{fig:frame_reconstruct1}, \ref{fig:frame_reconstruct2} and \ref{fig:frame_reconstruct3} show the visualization results of pixel reconstruction of masked patches. For each example in each figure, there is a textual sentence input to describe the sample frames. With a masking ratio of 50\%, our method can also effectively predict the original pixels of masked patches in the sampled frames, which can demonstrate the potential of adopting masked autoencoders for video-language pre-training.

\begin{table*}[t]
\vspace{-5mm}
\hspace{25mm}
\resizebox{0.7\textwidth}{!}{%
\centering
      \centering
        \begin{tabular}{p{4cm}<{\centering}|p{3.2cm}<{\centering}p{3.2cm}<{\centering}}
        config & pre-training & video QA\\
                    \midrule
            optimizer & \multicolumn{2}{c}{\multirow{1}{*}{AdamW~\cite{loshchilov2017decoupled}}} \\
            optimizer momentum & \multicolumn{2}{c}{\multirow{1}{*}{$\beta_{\text{1}}$=0.9, $\beta_{\text{2}}$=0.999}} \\
           learning rate schedule & \multicolumn{2}{c}{\multirow{1}{*}{cosine decay~\cite{loshchilov2016sgdr}}} \\
            weight decay &\multicolumn{2}{c}{\multirow{1}{*}{0.02}} \\
            image size & \multicolumn{2}{c}{\multirow{1}{*}{224$\times$224}} \\
            training epochs & \multicolumn{2}{c}{\multirow{1}{*}{10}} \\
            augmentation & \multicolumn{2}{c}{\multirow{1}{*}{random resize, horizontal flip, crop}}  \\
            base learning rate &  1e-4 & 1e-5\\
            min learning rate &  1e-5 & 1e-6\\
            batch size & 128 & 32 \\
            warmup epochs & 1 & 0.5\\
            inference frames & - & 12 \\
        \end{tabular}
}
        \caption{Pre-training and video question answering (video QA) configurations. Note that one single value is listed if all tasks share the same value, and the settings of all benchmarks for video QA are kept the same.}
          \label{table_be1}
\end{table*}

\begin{table*}[t]
\vspace{-60mm}
\hspace{12mm}
\resizebox{0.85\textwidth}{!}{%
\centering
      \centering
        \begin{tabular}{p{4cm}<{\centering}|p{3cm}<{\centering}p{3cm}<{\centering}p{3cm}<{\centering}}
        config & MSRVTT & DiDeMo & ActityNet Captions \\
            \midrule
            optimizer & \multicolumn{3}{c}{\multirow{1}{*}{AdamW~\cite{loshchilov2017decoupled}}} \\
            optimizer momentum & \multicolumn{3}{c}{\multirow{1}{*}{$\beta_{\text{1}}$=0.9, $\beta_{\text{2}}$=0.999}} \\
           learning rate schedule & \multicolumn{3}{c}{\multirow{1}{*}{cosine decay~\cite{loshchilov2016sgdr}}} \\
            base learning rate &  \multicolumn{3}{c}{\multirow{1}{*}{1e-5}} \\
            min learning rate &  \multicolumn{3}{c}{\multirow{1}{*}{1e-6}} \\
            weight decay & \multicolumn{3}{c}{\multirow{1}{*}{0.02}} \\
            image size & \multicolumn{3}{c}{\multirow{1}{*}{224$\times$224}} \\
            batch size & \multicolumn{3}{c}{\multirow{1}{*}{32}} \\
            augmentation & \multicolumn{3}{c}{\multirow{1}{*}{random resize, horizontal flip, crop}}  \\
            warmup epochs & \multicolumn{3}{c}{\multirow{1}{*}{0}} \\
            training epochs & 5 & 10 & 10 \\
            inference frames & 12 & 12 & 32 \\
            
        \end{tabular}
}
        \caption{Fine-tuning configurations for MSRVTT~\cite{xu2016msr}, DiDeMo~\cite{anne2017localizing} and ActivityNet Captions~\cite{krishna2017dense} benchmarks. Note that one single value is listed if all tasks share the same value.}
        \vspace{-3mm}
          \label{table_be2}
\end{table*}

\begin{figure*}[t]
  \hspace{-1mm}
\centering
\includegraphics[width=0.95\linewidth]{sup3.pdf}
    \caption{\textbf{The examples of pixel prediction for masked patches.} There're six examples totally, for each sample, the top row represents the original frames, the middle row means the masked frames, and the bottom row denotes the pixel reconstruction for masked patches.}
\label{fig:frame_reconstruct3}
 \vspace{-5mm}
\end{figure*}

\end{document}